%% file: main.tex
\documentclass[manuscript,screen]{acmart}

\AtBeginDocument{%
  }

\setcopyright{acmcopyright}
\copyrightyear{2022}
\acmYear{2022}
\acmDOI{XXXXXXX.XXXXXXX}

\acmPrice{15.00}
\acmISBN{978-1-4503-XXXX-X/18/06}

\usepackage{bm}
\usepackage{subcaption}

\usepackage{multirow}
\usepackage{mathtools,amssymb,euscript,yfonts,latexsym}
\usepackage[capitalise]{cleveref}
\usepackage{makecell}

\usepackage{color,colortbl}
\usepackage{textcomp}

\definecolor{LightCyan}{rgb}{0.8,1,1}
\definecolor{Gray}{gray}{0.9}

\renewcommand{\paragraph}[1]{\vspace{1.25mm}\noindent\textbf{#1}}

\newcommand{\E}{\mathbb{E}}

\makeatletter
\usepackage{xspace}
\def\@onedot{\ifx\@let@token.\else.\null\fi\xspace}
\DeclareRobustCommand\onedot{\futurelet\@let@token\@onedot}

\begin{document}



\title{A Complete Survey on Generative AI (AIGC): Is ChatGPT from GPT-4 to GPT-5 All You Need?}



\author{Chaoning Zhang}
\affiliation{%
  \institution{Kyung Hee University}
  \country{South Korea}
}
\email{chaoningzhang1990@gmail.com}

\author{Chenshuang Zhang}
\affiliation{%
  \institution{KAIST}
  \country{South Korea}
}
\email{zcs15@kaist.ac.kr}

\author{Sheng Zheng}
\affiliation{%
  \institution{Beijing Institute of Technology}
  \country{China}
}
\email{zszhx2021@gmail.com}

\author{Yu Qiao}
\affiliation{%
  \institution{Kyung Hee University}
  \country{South Korea}
}
\email{qiaoyu@khu.ac.kr}

\author{Chenghao Li}
\affiliation{%
  \institution{KAIST}
  \country{South Korea}
}
\email{lch17692405449@gmail.com}

\author{Mengchun Zhang}
\affiliation{%
  \institution{KAIST}
  \country{South Korea}
}
\email{zhangmengchun527@gmail.com}

\author{Sumit Kumar Dam}
\affiliation{%
  \institution{Kyung Hee University}
  \country{South Korea}
}
\email{skd160205@khu.ac.kr}

\author{Chu Myaet Thwal}
\affiliation{%
  \institution{Kyung Hee University}
  \country{South Korea}
}
\email{chumyaet@khu.ac.kr}

\author{Ye Lin Tun}
\affiliation{%
  \institution{Kyung Hee University}
  \country{South Korea}
}
\email{yelintun@khu.ac.kr}

\author{Le Luang Huy}
\affiliation{%
  \institution{Kyung Hee University}
  \country{South Korea}
}
\email{quanghuy69@khu.ac.kr}

\author{Donguk kim}
\affiliation{%
  \institution{Kyung Hee University}
  \country{South Korea}
}
\email{g9896@khu.ac.kr}

\author{Sung-Ho Bae}
\affiliation{%
  \institution{Kyung Hee University}
  \country{South Korea}
}
\email{shbae@khu.ac.kr}

\author{Lik-Hang Lee}
\affiliation{%
  \institution{Hong Kong Polytechnic University}
  \country{Hong Kong (China)}
}
\email{iskweon77@kaist.ac.kr}

\author{Yang Yang}
\affiliation{%
  \institution{University of Electronic Science and technology}
  \country{China}
}
\email{dlyyang@gmail.com}

\author{Heng Tao Shen}
\affiliation{%
  \institution{University of Electronic Science and technology}
  \country{China}
}
\email{shenhengtao@hotmail.com}

\author{In So Kweon}
\affiliation{%
  \institution{KAIST}
  \country{South Korea}
}
\email{iskweon77@kaist.ac.kr}

\author{Choong Seon Hong}
\affiliation{%
  \institution{Kyung Hee University}
  \country{South Korea}
}
\email{cshong@khu.ac.kr}

\renewcommand{\shortauthors}{Zhang et al.}


\input{main_AIGC_abstract}




\maketitle
{
  \hypersetup{linkcolor=black}
  \tableofcontents
}

\input{main_AIGC_text}


\bibliographystyle{ACM-Reference-Format}
\bibliography{bib_mixed,bib_local}



\end{document}

%% file: main_AIGC_abstract.tex
\begin{abstract}


As ChatGPT goes viral, generative AI (AIGC, a.k.a AI-generated content) has made headlines everywhere because of its ability to analyze and create text, images, and beyond. With such overwhelming media coverage, it is almost impossible for us to miss the opportunity to glimpse AIGC from a certain angle. In the era of AI transitioning from pure analysis to creation, it is worth noting that ChatGPT, with its most recent language model GPT-4, is just a tool out of numerous AIGC tasks . Impressed by the capability of the ChatGPT, many people are wondering about its limits: can GPT-5 (or other future GPT variants) help ChatGPT unify all AIGC tasks for diversified content creation? Toward answering this question, a comprehensive review of existing AIGC tasks is needed. As such, our work comes to fill this gap promptly by offering a first look at AIGC, ranging from its techniques to applications. Modern generative AI relies on various technical foundations, ranging from model architecture and self-supervised pretraining to generative modeling methods (like GAN and diffusion models). After introducing the fundamental techniques, this work focuses on the technological development of various AIGC tasks based on their output type, including text, images, videos, 3D content, etc., which depicts the full potential of ChatGPT's future. Moreover, we summarize their significant applications in some mainstream industries, such as education and creativity content. Finally, we discuss the challenges currently faced and present an outlook on how generative AI might evolve in the near future.


\end{abstract}

%% file: main_AIGC_text.tex
\section{Introduction}\label{sec:introduction}
Generative AI (AIGC, a.k.a AI-generated content) has made headlines with intriguing tools like ChatGPT or DALL-E~\cite{ramesh2021zero}, suggesting a new era of AI is coming. Under such overwhelming media coverage, the general public are offered many opportunities to have a glimpse of AIGC. However, the content in the media report tends to be biased or sometimes misleading. Moreover, impressed by the powerful capability of ChatGPT, many people are wondering about its limits. Very recently, OpenAI released GPT-4~\cite{OpenAI2023} which demonstrates remarkable performance improvement over the previous variant GPT-3 as well multimodal generation capability like understanding images. Impressed by the powerful capability of GPT-4 powered by AIGC, many are wondering about its limits: can GPT-5 (or other GPT variants) help next-generation ChatGPT unify all AIGC tasks? Therefore, a comprehensive review of generative AI serves as a groundwork to respond to the inevitable trend of AI-powered content creation. 
More importantly, our work comes to fill this gap in a timely manner. 

The goal of conventional AI is mainly to perform classification~\cite{lu2007survey} or regression~\cite{lewis2015applied}. Such a discriminative approach renders its role mainly for analyzing existing data. Therefore conventional AI is also often termed analytical AI. By contrast, generative AI differentiates by creating new content. However, generative AI often also requires the model to first understand some existing data (like text instruction) before generating new content~\cite{brown2020language,ramesh2022hierarchical}. From this perspective, analytical AI can be seen as the foundation of modern generative AI and the boundary between them is often ambiguous. Note that analytical AI tasks also generate content. For example, the label content is generated in image classification~\cite{krizhevsky2012imagenet}. Nonetheless, image recognition is often not considered in the category of generative AI because the label content has low dimensionality. Typical tasks for generative AI involve generating high-dimensional data, like text or images. Such generated content can also be used as synthetic data for alleviating the need for more data in deep learning~\cite{he2022synthetic}. An overview of the popularity of generative AI as well as its underlying reasons, is presented in Sec.\ref{sec:overview}.

As stated above, what distinguishes generative AI from conventional one lies in its generated content. With this said, generative AI is conceptually similar to AIGC (a.k.a. AI-generated content)~\cite{AIGC2022whitepaper}. In the context of describing AI-based content generation, these two terms are often interchangeable. In this work, we call the content generation tasks AIGC for simplicity. For example, ChatGPT is a tool for the AIGC task termed ChatBot~\cite{caldarini2022literature}, which is the tip of the iceberg considering the variety of AIGC tasks. Despite the high resemblance between generative AI and AIGC, these two terms have a nuanced difference. AIGC focuses on the tasks for content generation, while generative AI additionally considers the fundamental technical foundations that support the development of various AIGC tasks. In this work, we divide those underlying techniques into two classes. The first class refers to the generative modeling techniques, like GAN~\cite{goodfellow2014generative} and diffusion model~\cite{ho2020denoising}, which are directly related to generative AI for content creation. The second class of AI techniques mainly consists of backbone architecture (like Transformer~\cite{vaswani2017attention}) and self-supervised pretraining (like BERT~\cite{devlin2018bert} or MAE~\cite{he2022masked}). Some of them are developed in the context of analytical AI. However, they have also become essential for demonstrating competitive performance, especially in challenging AIGC tasks. Considering this, both classes of underlying techniques are summarized in Sec.\ref{sec:basictech}. 

On top of these basic techniques, numerous AIGC tasks have become possible and can be straightforwardly categorized based on the generated content type. The development of various AIGC tasks is summarized in Sec.\ref{sec:text}, Sec.\ref{sec:image} and Sec.\ref{sec:others}. Specifically, Sec.\ref{sec:text} and Sec.\ref{sec:image} focus on text output and image output, respectively.  For text generation, ChatBot~\cite{caldarini2022literature} and machine translation~\cite{yang2020survey} are two dominant tasks. Some text generation tasks also take other modalities as the input, for which we mainly focus on image and speech. For image generation, two dominant tasks are image restoration and editing~\cite{liu2022survey}. More recently, text-to-image has attracted significant attention. Beyond the above two dominant output types (\textit{i.e.} text and image), Sec.\ref{sec:others} covers other types of output, such as Video, 3D, Speech, etc. 

As technology advances, the AIGC performance gets satisfactory for more and more tasks. For example, ChatBot used to be limited to answering simple questions. However, the recent ChatGPT has been shown to understand jokes and generate code under simple instruction. Text-to-image used to be considered a challenging task; however, recent DALL-E 2~\cite{ramesh2022hierarchical} and stable diffusion~\cite{rombach2022high} have been able to generate photorealistic images. Therefore, opportunities of applying the AIGC to the industry emerge. Sec.\ref{sec:industry} covers the application of AIGC in various industries, including entertainment, digital art, media/advertising, education, etc. Along with the application of AIGC in the real world, numerous challenges like ethical concerns have also emerged and they are disused in Sec.\ref{sec:society}. Alongside the current challenges, an outlook on how generative AI might evolve is also presented. 

\begin{figure*}[t]
\centering
\includegraphics[width=1.0\textwidth]{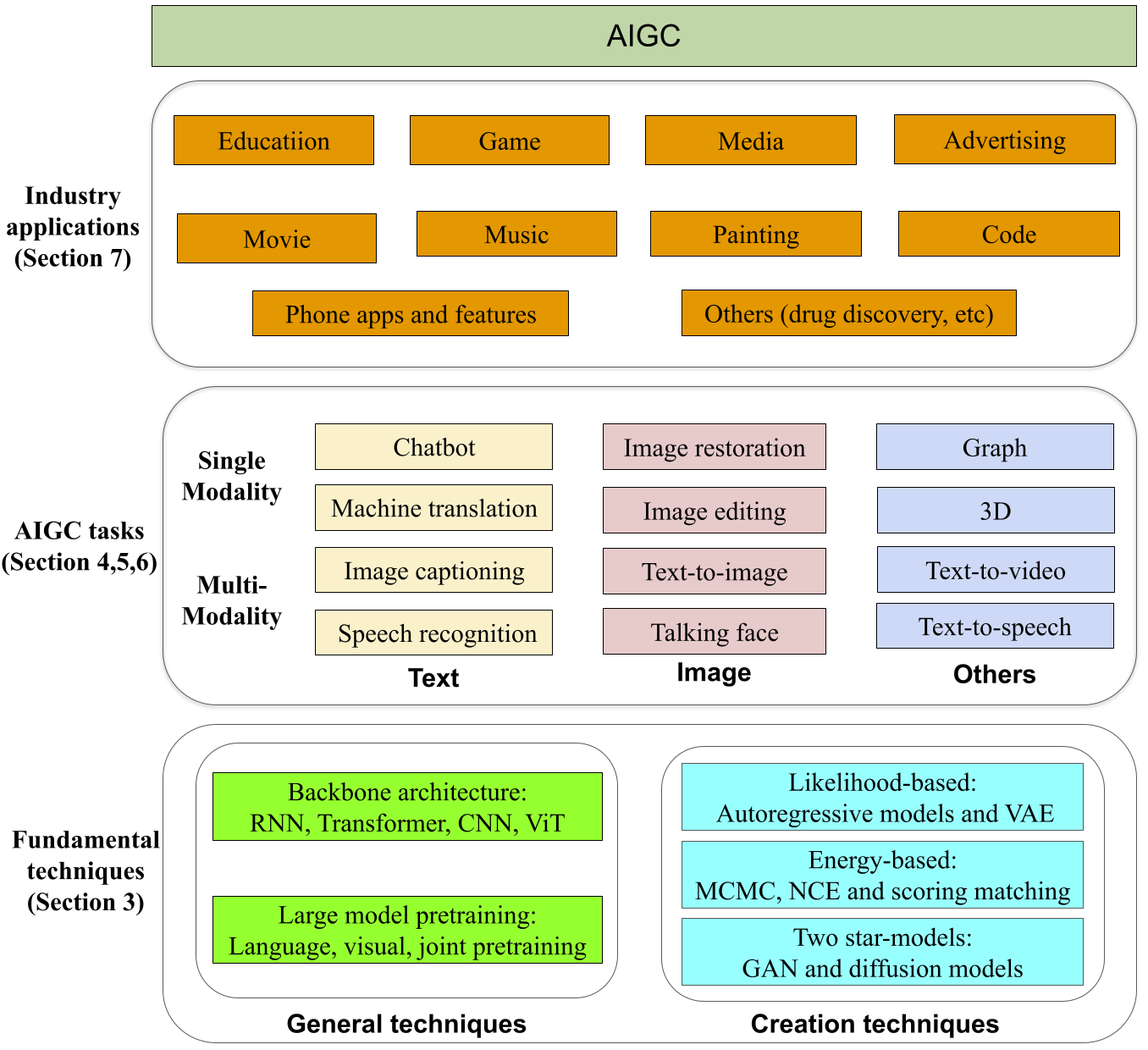}
\caption{An overview of generative AI (AIGC): fundamental techniques, core AIGC tasks, and industrial applications.}

\label{fig:aigc_tasks_overview}
\end{figure*}

Overall, this work conducts a survey on generative AI through the lens of generated content (\textit{i.e.} AIGC tasks), covering its underlying basic techniques, task-wise technological development, application in the industry as well as its social impact. An overview of the paper structure is presented in Figure  \ref{fig:aigc_tasks_overview}. 

\section{Overview} \label{sec:overview}
Adopting AI for content creation has a long history. IBM made the first public demonstration of a machine translation system at its head office in New York in 1954. The first computer-generated music came out with the name ``Illiac Suite" in 1957. Such early attempts and proof-of-concept successes caused a high expectation of the AI future, which motivated governments and companies to invest numerous resources in AI. Such a high boom in investment, however, did not yield the expected output. After that, a period called AI winter came, which dramatically undermines the development of AI and its applications. Entering the 2010s, AI has again become popular again, especially after the success of AlexNet~\cite{krizhevsky2012imagenet} for ImageNet classification in 2012. Entering the 2020s, AI has entered a new era of not only understanding existing data but also creating new content~\cite{brown2020language,ramesh2022hierarchical}. This section provides an overview of generative AI by focusing on its popularity and why it gets popular. 

\subsection{Popularity indicated by search interest}
A good indicator of `how popular a certain term is' refers to search interest. Google provides a promising tool to visualize search frequency, called Google trends. Although alternative search engines might provide similar functions, we adopt Google trends because Google is one of the most widely used search engines in the world.

\begin{figure*}[t]
    \centering   \includegraphics[width=0.45\linewidth]{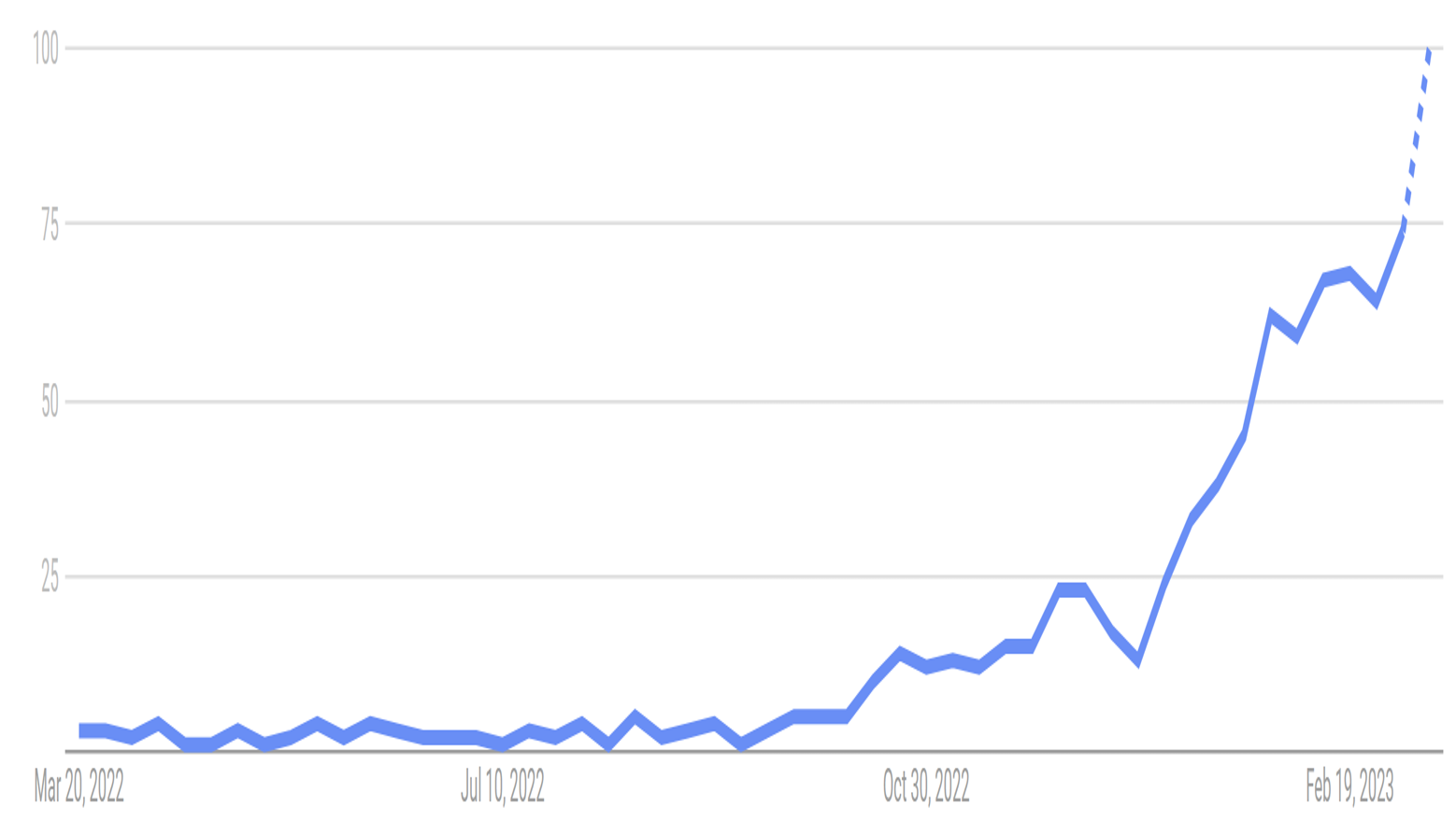}
    \includegraphics[width=0.45\linewidth]{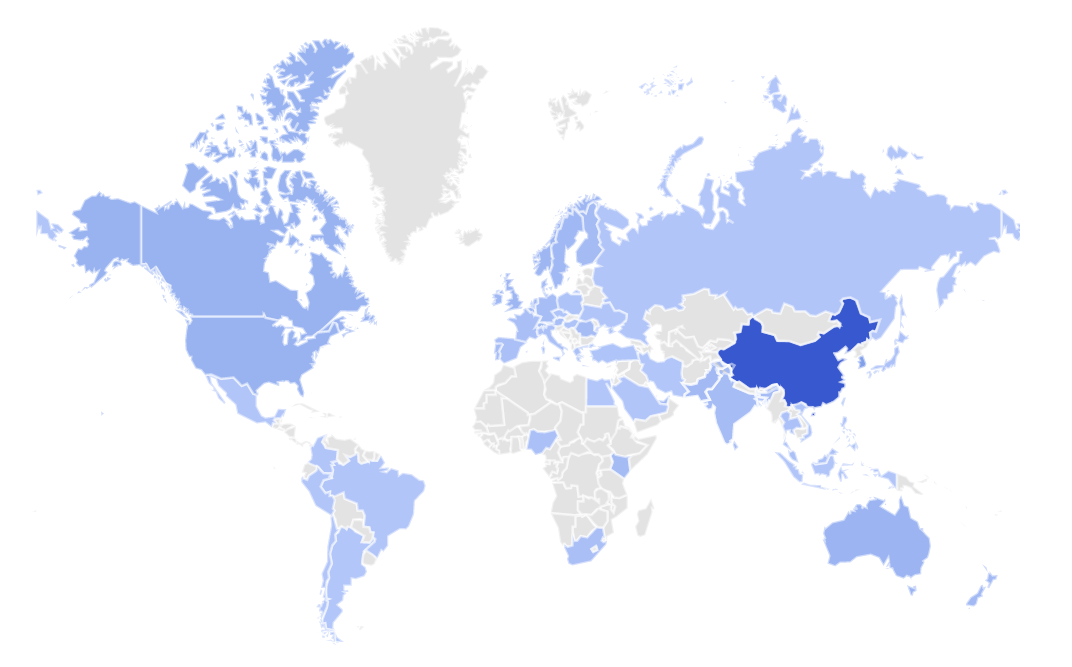}
    \caption{Search interest of generative AI: Timeline trend (left) and region-wise interest (right). The color darkness on the right part indicates the rank interest level.}
    \label{fig:GAI_trend}
\end{figure*}

\begin{figure*}[t]
    \centering   \includegraphics[width=0.45\linewidth]{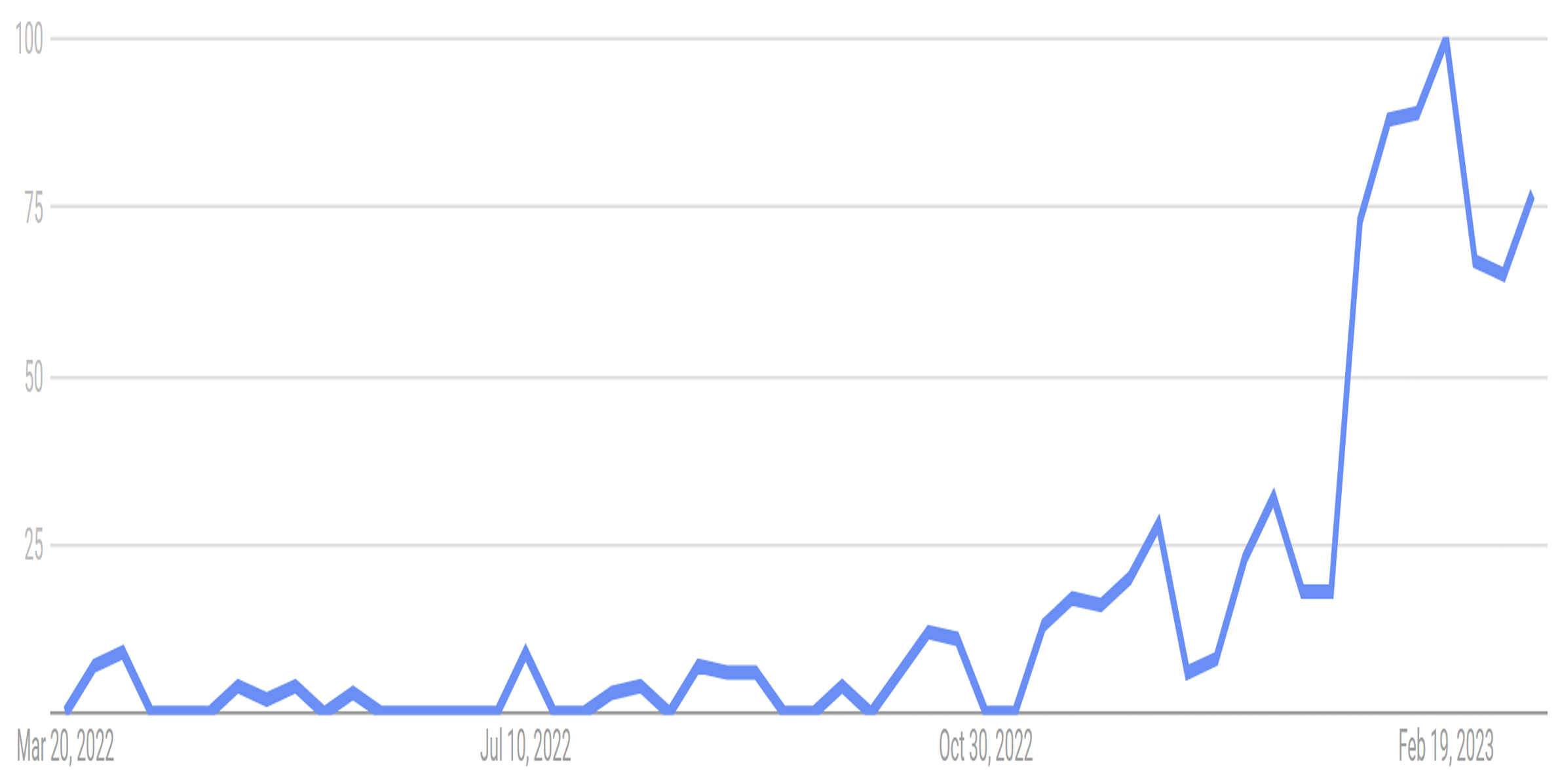}
    \includegraphics[width=0.45\linewidth]{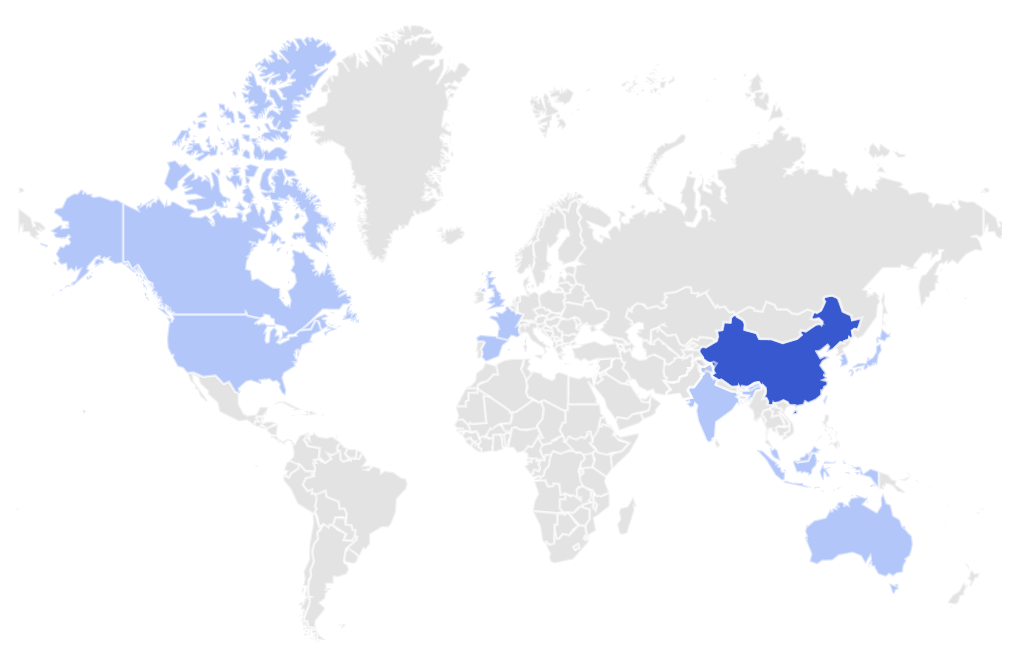}
    \caption{Search interest of AIGC: Timeline trend (left) and region-wise interest (right). The color darkness on the right part indicates the rank interest level.}
    \label{fig:AIGC_trend}
\end{figure*}

\textbf{Interest over time and by region.} Figure~\ref{fig:GAI_trend} (left) shows the search interest of generative AI, which indicates that the search interest significantly increased in the past year, especially after October 2022. Entering 2013, this search interest reaches a new height. A similar trend is observed for the term AIGC, see Figure~\ref{fig:AIGC_trend} (left). Except for interest over time, Google trends also provides region-wise search interest. The search heatmaps for generative AI and AIGC are shown in Figure~\ref{fig:GAI_trend} (right) and Figure~\ref{fig:AIGC_trend} (right), respectively. For both terms, the main hot regions include Asia, Northern America, and Western Europe. Most notably, for both terms, China ranks highest among all countries with a search interest of 100, followed by around 30 in Northern America and 20 in Western Europe. It is worth mentioning that some small but tech-oriented countries also have a very high search interest in generative AI. For example, the three countries that rank top on the country-wise search interest are Singapore (59), Israel (58), and South Korea (43).

\begin{figure*}[!htbp]
    \centering   \includegraphics[width=0.45\linewidth]{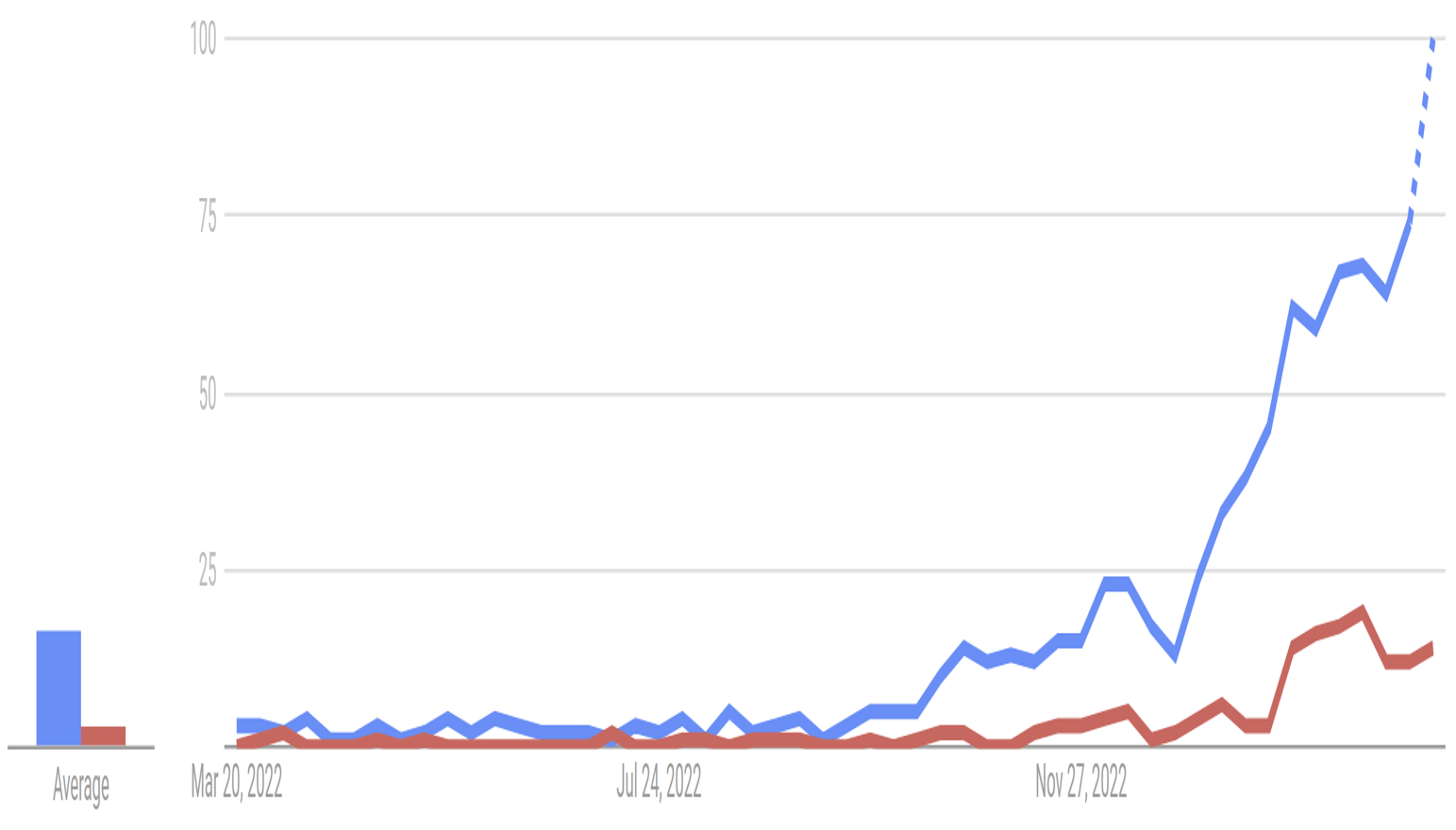}
    \includegraphics[width=0.45\linewidth]{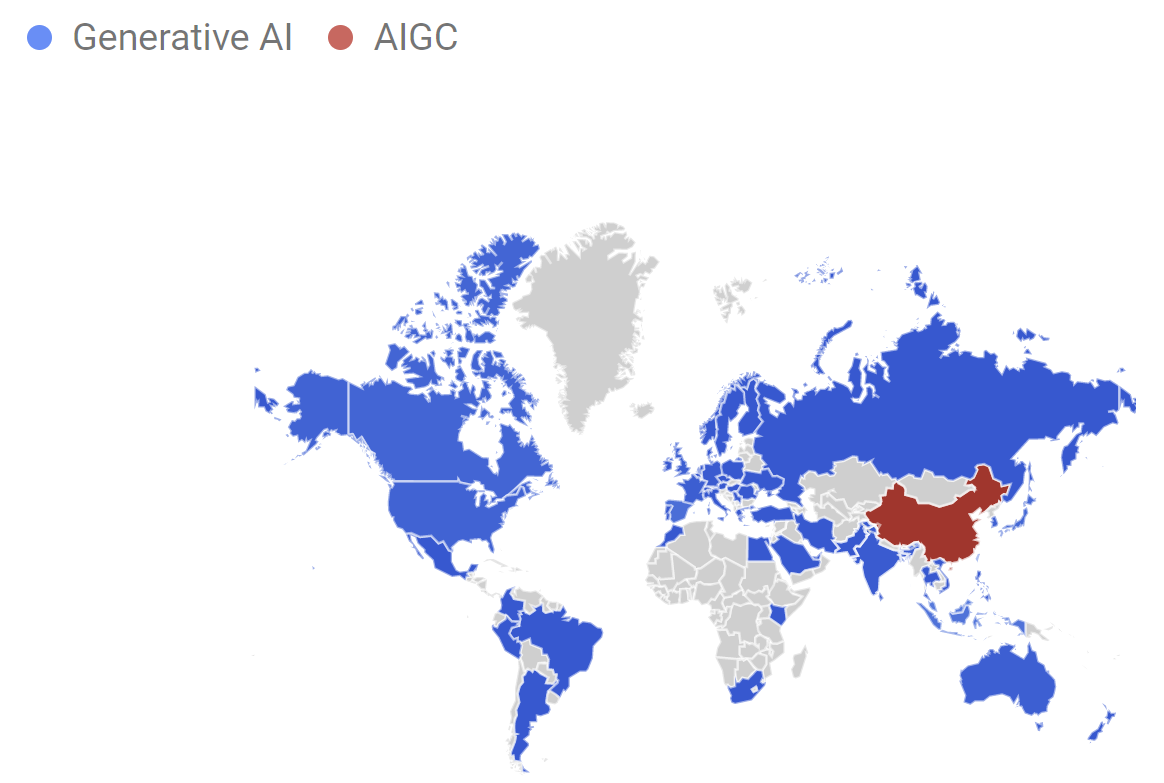}
    \caption{Search interest comparison between generative AI and AIGC: Timeline trend (left) and region-wise interest (right).}
    \label{fig:GAI_vs_AIGC}
\end{figure*}

\textbf{Generative AI \textit{v.s.} AIGC.} Figure~\ref{fig:GAI_vs_AIGC} shows a comparison between generative AI and AIGC for the search interest. Here, we define the interest ratio of generative AI and AIGC as GAI/AIGC. A major observation is that China prefers to use the term AIGC compared with generative AI with the GAI/AIGC ratio being 15/85. By contrast, the GAI/AIGC in the US is 90/10. In many countries, including Russia and Brazil, the  GAI/AIGC is 100/0. Overall, most countries prefer generative AI to AIGC, which makes generative AI have an overall higher search interest than AIGC. The reason that China becomes the leading country to adopt the term AIGC is not fully clear. A possible explanation is that AIGC is shortened to a single word and thus is easier to use. We also search the Chinese version of generative AI and AIGC on Google trends, however, the current demonstration is not sufficient. 

\subsection{Why does it get popular?}
The recent surging interest in generative AI in the last year can be mainly attributed to the emergence of intriguing tools like Stable diffusion or ChatGPT. Here, we discuss why generative AI gets popular by focusing on what factors contributed to the advent of such powerful AIGC tools. The reasons are summarized from two perspectives: content need and technology conditions. 
 
\subsubsection{Content need}
The way we communicate and interact with the world has been fundamentally changed by the Internet, for which \textit{digital content} plays a key role. Over the last few decades, the content on the web has also undergone multiple major changes. In the Web 1.0 era (the 1990s-2004), the Internet was primarily used to access and share information, with websites mainly static. There was little interaction between users and the primary mode of communication was one-way, with users accessing information but not contributing or sharing their own content. The content was largely text-based and it was mainly generated by professionals in the relative fields, like journalists generating news articles. Therefore, such content is often called Professional Generated Content (PGC), which has been dominated by another type of content, termed User Generated Content (UGC)~\cite{paulussen2008user,koolen2013social,timoshenko2019identifying}. In contrast to PGC, UGC in Web 2.0~\cite{o2009web} is mainly generated by users on social media, like Facebook~\cite{kim2016power}, Twitter~\cite{liu2017investigation}, Youtube~\cite{holland2016youtube}, etc. Compared with PGC, the volume of UGC is significantly larger, however, its quality might be inferior. 

We are currently transitioning from Web 2.0 to Web 3.0~\cite{rudman2016defining}. With defining features of being decentralized and intermediary-free, Web 3.0 also relies on a new content generation type beyond PGC and UGC to address the trade-off between volume and quality. AI is widely recognized as a promising tool for addressing this trade-off. For example, in the past, only those users that have a long period of practice could draw images of decent quality. With text-to-image tools (like stable diffusion~\cite{rombach2022high}), anyone can create drawing images with a plain text description. Such a combination of user imagination power and AI execution power makes it possible to generate new types of images at an unprecedented speed. Beyond image generation, AIGC tasks also facilitate generating other types of content. 

Another change AIGC brings is that the boundary between content consumer and creator becomes vague. In Web 2.0, Content generators and consumers are often different users. With AIGC in Web 3.0, however, data consumers are now able to become data creators, as they are able to use AI algorithms and technology to generate their own original content, and it allows them to have more control over the content they produce and consume, making them use their own data and AI technology to produce content that is tailored to their specific needs and interests. Overall, the shift towards AIGC has the potential to greatly transform the way data is consumed and produced, giving individuals and organizations more control and flexibility in the content they create and consume. In the following, we discuss why AIGC has become popular now. 

\subsubsection{Technology conditions}
When it comes to AIGC technology, the first thing that comes into mind is often machine (deep) learning algorithm, while overlooking its two important conditions: data access and compute resources. 

\textbf{Advances in data access.} Deep learning refers to the practice of training a model on data. The model performance heavily relies on the size of the training data. Typically, the model performance increases with more training samples. Taking image classification as an example, ImageNet~\cite{deng2009imagenet} with more than 1 million images is a commonly used dataset for training the model and validating the performance. Generative AI often requires an even larger dataset, especially for challenging AIGC tasks like text-to-image. For example, approximately 250M images were used for training DALL-E~\cite{ramesh2021zero}. DALL-E 2~\cite{ramesh2022hierarchical}, on the other hand, used approximately 650M images. ChatGPT was built on top of GPT3~\cite{brown2020language} partly trained on CommonCrawl dataset, which has 45TB of compressed plaintext before filtering
and 570GB after filtering. Other datasets like WebText2, Books1/2, and Wikipedia are involved in the training of GPT3. Accessing such a huge dataset becomes possible mainly due to the Internet. 

\textbf{Advances in computing resources.} Another important factor contributing to this development of AIGC is advanced in computing resources. Early AI algorithm was run on CPU, which cannot meet the need of training large deep learning models. For example, AlexNet~\cite{krizhevsky2012imagenet} was the first model trained on full ImageNet and the training was done on Graphics Processing Units (GPUs). GPUs were originally designed for rendering graphics in video games but have become increasingly common in deep learning. GPUs are highly parallelized and can perform matrix operations much faster than CPUs. Nvidia is a leading company in manufacturing GPUs. The computing capability of its CUDA has improved from the first CUDA-capable GPU (GeForce 8800) in 2006 to the recent GPU (Hopper) with hundreds of times more computing power. The price of GPUs can range from a few hundred dollars to several thousand dollars, depending on the number of cores and memory. Tensor Processing Units (TPUs) are specialized processors designed by Google specifically for accelerating neural network training. TPUs are available on the Google Cloud Platform, and the pricing varies depending on usage and configuration. Overall, the price of computing resources is on the trend of becoming more affordable. 

\section{Fundamental techniques behind AIGC} \label{sec:basictech}

In this work, we perceive AIGC as a set of tasks or applications that generates content with AI methods. Before introducing AIGC, we first visit the fundamental techniques behind AIGC, which fall in the scope of generative AI at the technical level. Here, we summarize the fundamental techniques by roughly dividing them into two classes: Generative techniques and Creation techniques. Specifically, Creation techniques refer to the techniques that are able to generate various contents, e.g., GAN and diffusion model. Meanwhile, General techniques cannot generate content directly but are essential for the development of AIGC, e.g., the Transformer architecture. In this section, we provide a brief summary of the required techniques for AIGC.

\subsection{General techniques in AI}
After the phenomenal success of AlexNet~\cite{krizhevsky2012imagenet}, there is a surging interest in deep learning, which somewhat becomes a synonym for AI. In contrast to traditional rule-based algorithms, deep learning is a data-driven method that optimizes the model parameters with a stochastic gradient. The success of deep learning in obtaining a superior feature representation depends on better backbone architecture and more data, which greatly accelerates the development of AIGC.

\subsubsection{Backbone architecture}
As two mainstream fields in deep learning, the research on
natural language processing (NLP) and computer vision (CV) have significantly improved the backbone architectures and inspired various applications of improved backbones in other fields, e.g., the speech area. In the NLP field, Transformer~\cite{vaswani2017attention} has replaced recurrent neural networks (RNN) ~\cite{mikolov2010recurrent,medsker2001recurrent} to be the de-facto standard backbone. In the CV area, vision Transformer (ViT)~\cite{dosovitskiy2020image} has also shown its power besides the traditional convolutional neural networks (CNN). Here, we will briefly introduce how these mainstream backbones work and their representative variants. 

\textbf{RNN architecture.} RNN is mainly adopted for handling data with time sequences, like language or audio. A vanilla RNN has three layers: input, hidden, and output. The information flow in RNN is in two directions. The first direction is from the input to the hidden layer and then to the output. 
What captures the \textit{recurrent} nature of RNN lies in its second information flow in the time direction. Except for the corresponding input, the current hidden state depends at time $t$ depends on the hidden state at time $t-1$. This two-flow design well handles the sequence order but suffers from exploding or vanishing gradients when the sequence gets long. To mitigate long-term dependency,
LSTM~\cite{hochreiter1997long} was introduced with a cell state that acts like a freeway to facilitate the information flow in the sequence direction. LSTM is one of the most popular methods for alleviating the gradient vanishing/exploding issue. With three types of gates, however, LSTM suffers from high complexity and a higher memory requirement. Gated Recurrent Unit (GRU)~\cite{chung2014empirical} simplifies LSTM by merging its cell and hidden states and replacing the forget and input gates with a so-called update state. Unitary RNN~\cite{arjovsky2016unitary} handles the gradient issue by implementing unitary matrices. Gated Orthogonal Recurrent Unit~\cite{jing2019gated} leverages the merits of both gate and unitary matrices. 
Bidirectional RNN~\cite{schuster1997bidirectional} improves vanilla RNN by capturing both past and future information in the cell, i.e., the state at time $t$ is calculated based on both time $t-1$ and $t+1$. Depending on the tasks, RNN can have various architectures with a different number of inputs and outputs: one-to-one, many-to-one, one-to-many, and many-to-many. The many-to-many can be used in machine translation and is also called the sequence-to-sequence (seq2seq) model~\cite{sutskever2014sequence}. Attention was introduced in~\cite{bahdanau2014neural} to make the model decoder see every encoder token and automatically decide the weights on them based on their importance.   

\begin{figure*}[!htbp]
    \centering
    \includegraphics[width=0.45\linewidth]{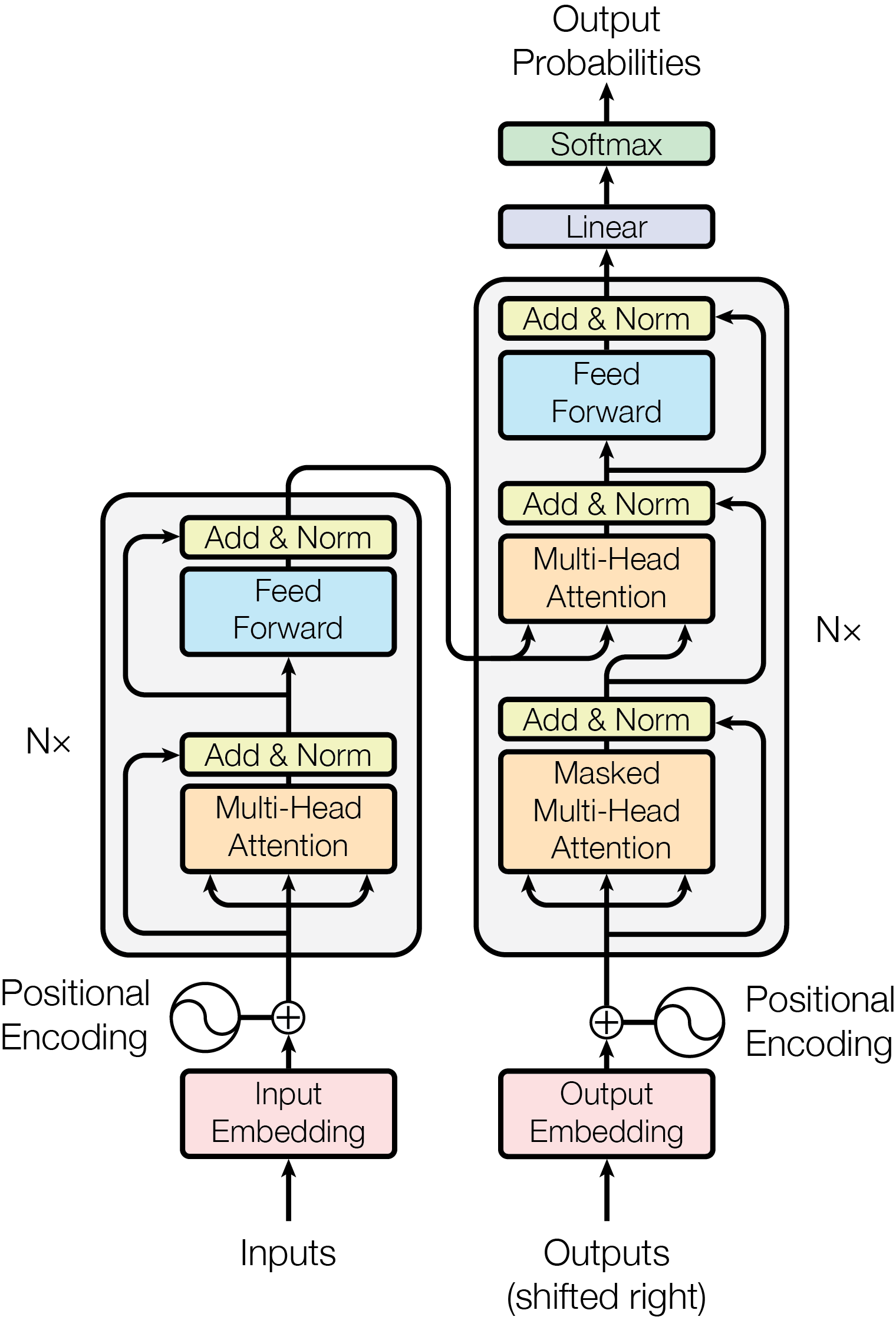}
    \caption{Transformer structure (figure obtained from \cite{vaswani2017attention}).}
    \label{fig:transformer}
\end{figure*}

\textbf{Transformer.} Different from Seq2seq with attention~\cite{bahdanau2014neural,luong2015effective,parikh2016decomposable}, a new variant of architecture discards the seq-2seq architecture and claims that attention is all you need~\cite{vaswani2017attention}. Such attention is called self-attention, and the proposed architecture is termed \textit{Transformer}~\cite{vaswani2017attention} (see Figure \ref{fig:transformer}). A standard Transformer consists of an encoder and a decoder and is developed based on residual connection~\cite{he2016deep} and layer normalization~\cite{ba2016layer}. Except for the Add $\&$ Norm module, the Transformer has two core components: multi-head attention and feed-forward neural network (a.k.a. MLP). The attention module adopts a multi-head design with the self-attention in the form of scaled dot-product defined as:
\begin{equation}
Attention(Q, K, V) = softmax(\frac{QK^T}{\sqrt{d_k}.})V
\end{equation}
Unlike RNNs, which build positional information by sequentially inputting sentence information, Transformer obtains powerful modeling capabilities by constructing global dependencies but also loses information with positional bias. Therefore, positional encoding is needed to enable the model to sense the positional information of the input signal. There are two types of positional encoding. Fixed position coding is represented by sinusoids and cosines of different frequencies. The learnable position encoding is composed of a set of learnable parameters. Transformer has become the de-facto standard method in NLP tasks. 

\textbf{CNN architecture.} After introducing RNN and Transformer in NLP field, we start to visit two mainstream backbones in CV area, i.e., CNN and ViT. 
CNNs have become a standard backbone in the field of computer vision. The core of CNN lies in its convolution layer. The convolution kernel (also known as filter) in the convolution layer is a set of shared weight parameters for operating on images, which is inspired by the biological visual cortex cells. The convolution kernel slides on the image and performs correlation operations with the pixel values on the image, finally obtaining the feature map and realizing the feature extraction of the image. GoogleNet\cite{szegedy2015going}, with its Inception module allowing multiple convolutional filter sizes to be chosen in each block, increased the diversity of convolutional kernels, thus the performance of CNN was improved. ResNet\cite{he2016deep} was a milestone for CNNs, introducing residual connections that stabilized training and enabled the models to achieve better performance through deeper modeling. After that, it became part of the binding in CNNs. In order to expand the work of ResNet, DenseNet\cite{huang2017densely} establishes dense connections between all the previous layers and the subsequent layers, thus enabling the model to have better modeling ability. EfficientNet\cite{tan2019efficientnet} uses a scaling method which uses a set of fixed scaling coefficients to uniformly scale the width, depth, and resolution of the convolutional neural network architecture, thus making the model more efficient.

\begin{figure}[!htbp]
    \centering
    \includegraphics[width=1.0\linewidth]{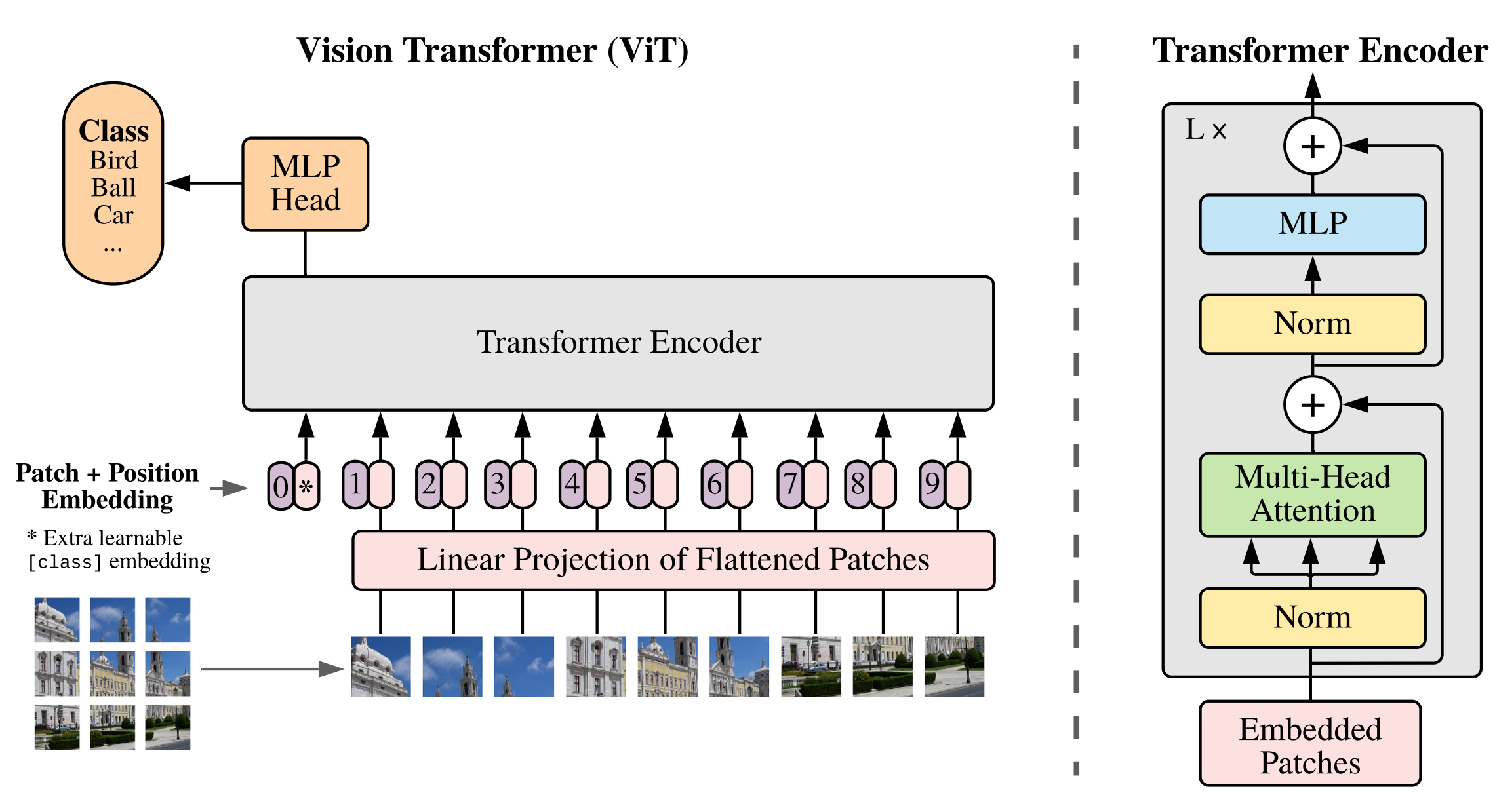}
    \caption{ViT structure (figure obtained from \cite{dosovitskiy2020image}).}
    \label{fig:ViT}
\end{figure}

\textbf{ViT architecture.} Inspired by the success of Transformer in NLP, numerous works have tried to apply Transformer to the field of CV with ViT\cite{dosovitskiy2020image} (see Figure \ref{fig:ViT}), being the first of its kind. ViT first flattens the image into a sequence of 2D patches and inserts a class token at the beginning of the sequence to extract classification information. After the embedding position encoding, the token embeddings are fed into a standard Transformer. This simple and effective implementation of ViT makes it highly scalable. Swin~\cite{liu2021swin} efficiently deals with image classification and dense recognition tasks by constructing hierarchical feature maps by merging image blocks at a deeper level, and due to its computation of self-attention only within each local window, it reduces computational complexity. DeiT\cite{touvron2020training} uses the teacher-student strategy for training, reducing the dependence of Transformer models on large data, by introducing distillation tokens. CaiT\cite{touvron2021going} introduces class attention to effectively increase the depth of the model. T2T\cite{yuan2021tokens} effectively localizes the model by Token Fusion and introduces hierarchical deep and narrow structures through the prior of CNNs by recursively aggregating adjacent Tokens into one Token. Through permutation equivariance, Transformers have liberated CNNs from their translation invariance, allowing for long-range dependencies and less inductive bias, making them more powerful modeling tools and better transferable to downstream tasks than CNNs. In the current paradigm of large models and large datasets, Transformers have gradually replaced CNNs as the mainstream model in the field of computer vision.

\subsubsection{Self-supervised pretraining}
Parallel to better backbone architecture, deep learning also benefits from self-supervised pertaining which can exploit a larger (unlabeled) training dataset. Here, we summarize the most relevant pretraining techniques to AIGC, and categorize them according to the training data type (e.g., language, vision, and joint pretraining).

\begin{figure}[!htbp]
    \centering
    \includegraphics[width=\linewidth]{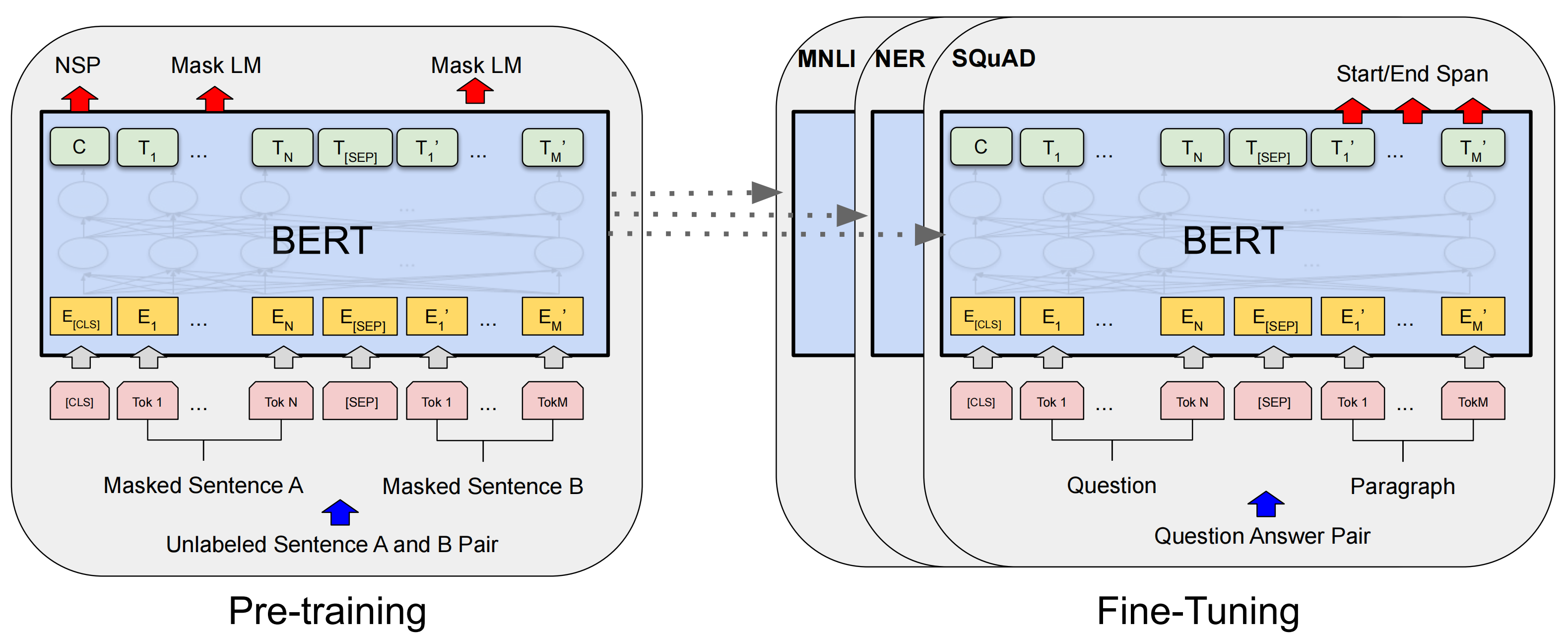}
    \caption{BERT structure (figure obtained from \cite{devlin2018bert}).}
    \label{fig:BERT}
\end{figure}

\textbf{Language pretraining.} There are three major types of language pretraining methods. The first type pretrains an encoder with masking, for which the representative work is BERT~\cite{devlin2018bert} (see Figure \ref{fig:BERT}). Specifically, BERT predicts the masked language tokens from the unmasked tokens. There is a significant discrepancy between the mask-then-predict pertaining task and downstream tasks, therefore masked language modeling like BERT is rarely used for text generation without finetuning. By contrast, autoregressive language pretraining methods are suitable for few-shot or zero-shot text generation. GPT family~\cite{radford2018improving,radford2019language,brown2020language} is the most popular one which adopts a decoder instead of an encoder. Specifically, GPT-1
~\cite{radford2018improving} is the first of its kind with GPT-2~\cite{radford2019language} and GPT-3~\cite{brown2020language} further investigating the role of massive data and large model in the transfer capacity. Based on GPT-3, the unprecedented success of ChatGPT has attracted great attention recently. Moreover, a stream of language models adopts both an encoder and decoder as the original Transformer. BART~\cite{lewis2019bart} perturbed the input with various types of noise and predicted the original clean input, like a denoising autoencoder. MASS~\cite{song2019mass} and PropheNet~\cite{qi2020prophetnet} follow BERT to take a masked sequence as the input of the encoder with the decoder predicting the masked tokens in an autoregressive manner. T5~\cite{raffel2020exploring} replaces the masked tokens with some random tokens. 

\begin{figure}[!htbp]
    \centering
    \includegraphics[width=0.8\linewidth]{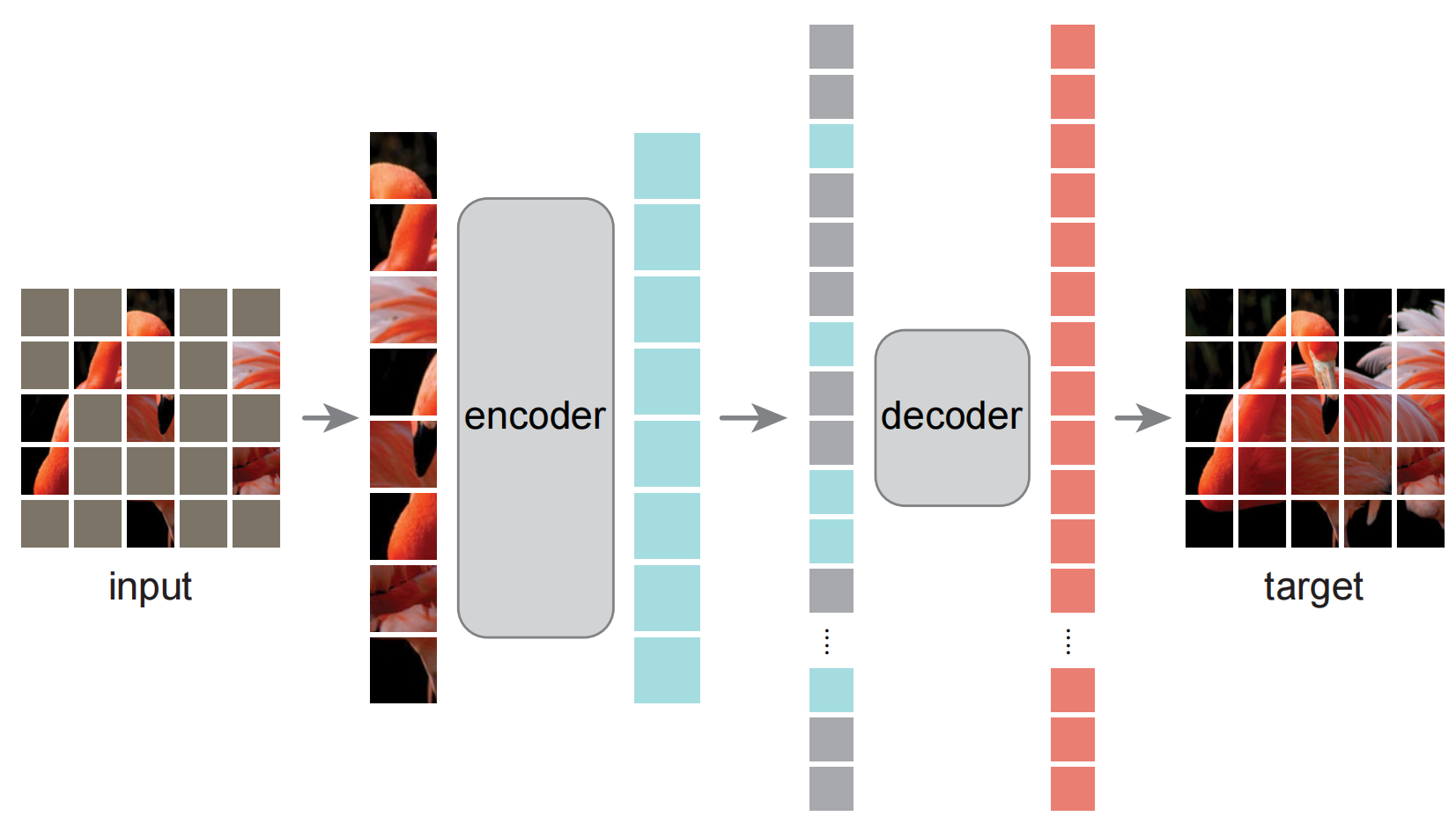}
    \caption{MAE structure (figure obtained from \cite{he2022masked}).}
    \label{fig:MAE}
\end{figure}

\textbf{Visual pretraining.}
To learn better representations of vision data during pretraining, self-supervised learning (SSL) has  been widely applied, and we term it  visual SSL. Visual SSL has undergone three stages. Early works focused on designing various pretext tasks like jigsaw puzzles~\cite{noroozi2016unsupervised} or predicting rotation~\cite{gidaris2018unsupervised}. Such pretraining yields better performance on the downstream task than training from scratch, which motivates contrastive learning methods~\cite{chen2020simple,he2020momentum,zhang2022dual}. Contrastive learning adopts joint embedding to minimize the representation distance between augmented images for learning augmentation-invariant representation. The representation in pure joint embedding can collapse to a constant regardless of the inputs, for which contrastive learning simultaneously maximizes the representation distance from negative samples. Negative-free joint-embedding methods have also been investigated in SimSiam~\cite{chen2021exploring} and BYOL~\cite{grill2020bootstrap}. How SimSiam works without negative samples have been investigated in~\cite{zhang2022how}. Inspired by the success of BERT in NLP for pertaining, BEiT~\cite{bao2022beit} applied masking modeling in vision and its success relies on a pre-trained VAE to obtain the visual token. Masked autoencoder (MAE)~\cite{he2022masked} (see Figure \ref{fig:MAE}) simplifies it to an end-to-end denoising framework by predicting the masked patches from the unmasked patches. Outperforming contrastive learning and negative-free joint-embedding methods, MAE has become a new variant of the visual SSL framework. Interested readers can refer~\cite{zhang2022survey} for more details.

\begin{figure}[t]
    \centering
    \includegraphics[width=\linewidth]{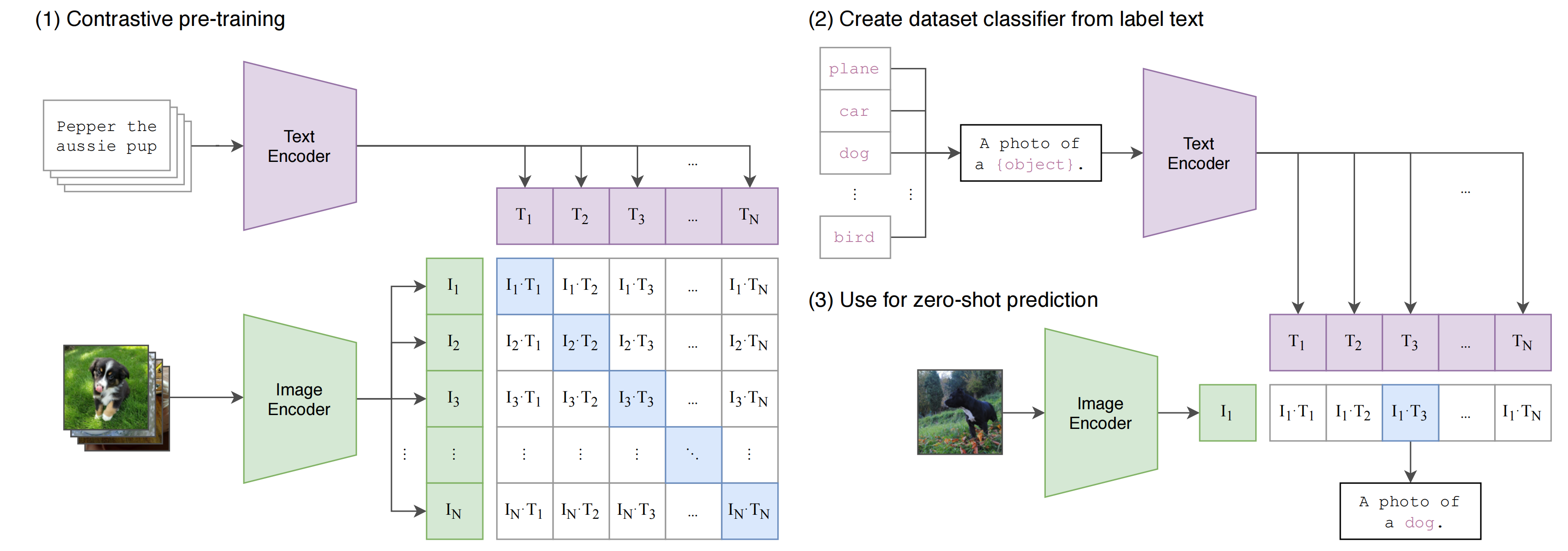}
    \caption{CLIP structure (figure obtained from \cite{radford2021learning}).}
    \label{fig:CLIP}
\end{figure}

\textbf{Joint pretraining.} With large datasets of image-text pairs collected from the Internet, multimodal learning~\cite{baltruvsaitis2018multimodal,xu2022multimodal} has made unprecedented progress to learn data representations, at the front of which is cross-modal matching~\cite{gan2022vision}. Contrastive pretraining is widely used to match the image embedding and text encoding in the same representation space~\cite{radford2021learning,jia2021scaling,yuan2021florence}. CLIP~\cite{radford2021learning} (see Figure \ref{fig:CLIP} is a pioneering work in this direction and is used in numerous text-to-image models, such as DALL-E 2~\cite{ramesh2022hierarchical}, Upainting~\cite{li2022upainting}, DiffusionCLIP~\cite{kim2021diffusionclip}. ALIGN~\cite{jia2021scaling} extended CLIP with noisy text supervision so that the text-image dataset requires no cleaning and can be scaled to a much larger size (from 400M to 1.8B). Florence~\cite{yuan2021florence} further expands the cross-modal shared representation from coarse scene to dine object and from static images to dynamic videos, etc. Therefore, the learned shared representation is more universal and shows superior performance~\cite{yuan2021florence}. 

\subsection{Creation techniques in AI}
Deep generative models (DGMs) are a group of probabilistic models that use neural networks to generate samples. Early attempts at generative modeling focused on pre-training with an autoencoder~\cite{rumelhart1985learning,ballard1987modular,hinton1993autoencoders}. A variant of autoencoder with masking has emerged to become a dominant self-supervised learning framework, and interested readers are encouraged to check a survey on masked autoencoder~\cite{zhang2022survey}. Unless specified, the use cases of deep generative models in this survey only consider generating new data. The generated data is typically high-dimensional, and therefore, predicting a label of a sample is not considered discriminative instead of generative modeling even though something like a label is also technically generated. 

Numerous DGMs have emerged and can be categorized into two major groups: likelihood-based and energy-based. Likelihood-based probabilistic models, like autoregressive models~\cite{graves2013generating} and flow models~\cite{dinh2014nice}, have a tractable likelihood which provides a straightforward method to optimize the model weights w.r.t. the log-likelihood of the observed (training) data. The likelihood is not fully tractable in variational autoencoders (VAEs)~\cite{kingma2013auto}, but a tractable lower bound can be optimized, thus VAE is also considered to lie in the likelihood-based group which specifies a normalized probability. By contrast, energy-based models~\cite{grenander1994representations,hinton2002training} are featured by the unnormalized probability, a.k.a. energy function. Without the constraint on the tractability of the normalizing constant, energy-based models are more flexible in parameterizing but difficult to train~\cite{song2021train}. Notably, GAN and diffusion models are highly related to energy-based models even though are developed from different motivations. In the following, we present an introduction to each class of likelihood-based models, followed by how the energy-based models can be trained as well as the mechanism behind GAN and diffusion models. 

\subsubsection{Likelihood-based models}
\textbf{Autoregressive models.} Autoregressive models learn the joint distribution of sequential data and predict each variable in the sequence with  previous time-step variables as inputs.  As shown in Eq.~\ref{eq:autoregressive}, autoregressive models assumes that the joint  distribution $p_\theta(x)$ can be decomposed to a product of conditional distributions.
\begin{align}
p_\theta(x) = p_\theta(x_1)p_\theta(x_2|x_1)...p_\theta(x_n|x_1,x_2,...,x_{n-1}), \label{eq:autoregressive}
\end{align}  
Although both rely on previous timesteps, autoregressive  models differ from  RNN architecture since the previous timesteps are given to the model as input instead of hidden states in RNN. In other words, 
autoregressive models can be seen as a feed-forward network that takes all the previous time-step variables as inputs. 
Early works model discrete data  with different functions estimating the conditional distribution, e.g. logistic regression in Fully Visible Sigmoid Belief Network (FVSBN)~\cite{gan2015learning} and  one hidden layer neural networks in  Neural Autoregressive Distribution Estimation (NADE) ~\cite{larochelle2011neural}. The following research  further extends to model the continuous variables~\cite{uria2013rnade,uria2016neural}. Autoregressive methods have been widely applied in multiple  areas, including computer vision (PixelCNN~\cite{van2016conditional} and PixelCNN++~\cite{salimans2017pixelcnn}), audio generation (WaveNet~\cite{oord2016wavenet}), natural language processing (Transformer~\cite{vaswani2017attention}).

\textbf{VAE.} Autoencoders are a family of models that first map the input to a low-dimension latent layer with an encoder and then reconstruct the input with a decoder. The entire encoder-decoder process  aims to learn the underlying data patterns and generate unseen samples~\cite{oussidi2018deep}. Variational autoencoder (VAE)~\cite{kingma2013auto} is an autoencoder that learns the data distribution $p(x)$ from latent space z, i.e., $p(x)=p(x|z)p(z)$, where $p(x|z)$ is learned by the decoder. In order to obtain $p(z)$, VAE~\cite{kingma2013auto} adopts Bayes' theorem and  approximates the posterior distribution $p(z|x)$  by the encoder. The VAE model is optimized toward a likelihood goal with regularizer~\cite{altosaar_jaan_2016_4462916}.

\subsubsection{Energy-based models}
With a tractable likelihood, autoregressive models and flow models allow a straightforward optimization of the parameters w.r.t. the log-likelihood of the data. This forces the model to be constrained in a certain form. For example, the autoregressive model needs to be factorized as a product of conditional probabilities, and the flow model must adopt invertible transformation. 

Energy-baed models specify probability up to an unknown normalizing constant, therefore, they are also known as non-normalzied probabilistic models. Without losing generality by assuming the energy-based model is over a single variable $\bm{x}$, we denote its energy as $E_{\theta}(\bm{x})$. Its probability density is then calculated as
\begin{align}
     p_{\theta}(\bm{x}) = \frac{\exp(- E_{\theta}(\bm{x}))}{z_{\theta}}
\label{eq:ebm}
\end{align}
where $z_{\theta}$ is the so-called normalizing constant and defined as $z_{\theta} = \int \exp(-E_{\theta}(\bm{x}))~\text{d}\bm{x}$. $z_{\theta}$ is an intractable integral, making optimizing energy-based models a challenging task.

\textbf{MCMC and NCE.} Early attempts at optimizing energy-based models opt to estimate the gradient of the log-likelihood with Markov chain Monte Carlo (MCMC) approaches, which require a cumbersome drawing of random samples. Therefore, some works aim to improve the efficiency of MCMC a representative work Langevin MCMC~\cite{parisi1981correlation,grenander1994representations}. Nonetheless, performing MCMCM to obtain requires large computation and contrastive divergence (CD)~\cite{hinton2002training} is a popular method to reduce the computation via approximation with various variants: persistent CD~\cite{tieleman2008training}, mean
field CD~\cite{welling2002new}, and multi-grid CD~\cite{gao2018learning}. Another line of work optimizes energy-based models via notice contrastive estimation (NCE)~\cite{gutmann2010noise}, which contrasts the probabilistic model with another noise distribution. 
Specifically, it optimizes the following loss:
\begin{align}
    \E_{p_d} \bigg[ \ln \frac{p_\theta(\bm{x})}{p_\theta(\bm{x}) + q_\phi(\bm{x})} \bigg] + \E_{q_\phi} \bigg[ \ln \frac{q_\phi(\bm{x})}{p_\theta(\bm{x}) + q_\phi(\bm{x})} \bigg], \label{eqn:noise-contrastive-divergence}
\end{align}

\textbf{Score matching.} For optimizing energy-based models, another popular MCMC-free method minimizes the derivatives of log probability density between the model and the observed data. The first-order of a log probability density function is called \textit{score} of the distribution ($s(\bm{x}) = \nabla_{\bm{x}} \text{log} p(\bm{x})$), therefore, this method is often termed \textit{score matching}. 
Unfortunately, the data score function $ s_d(\bm{x})$ is unavailable. Various attempts~\cite{vincent2011connection,saremi2018deep,song2019generative,pang2020efficient,shi2018spectral,song2020sliced} have been made to mitigate this issue, with a representative method called denoising score matching~\cite{vincent2011connection}. Denoising score matching approximates the score of data with noisy samples. 
The model takes a noisy sample as the input and predicts its noise. Therefore, it can be used for sampling clean samples from noise by iterative removing the noise~\cite{saremi2018deep,song2019generative}. 

\subsubsection{Two star-models: from GAN to diffusion model}
When it comes to deep generative models, what first comes to your mind? The answer depends on your background, however, GAN is definitely one of the most mentioned models. GAN stands for generative adversarial network~\cite{goodfellow2014generative} which was first proposed by Ian J. Goodfellow and his team in 2014 and rated as ``the most interesting idea in the last 10 years in machine learning" by Yann Lecun in 2016. As the pioneering work to generate images of reasonably high quality, GAN has been widely regarded as a de facto standard model for the challenging task of image synthesis. This long-time dominance has been recently challenged by a new family of deep generative models termed diffusion models~\cite{ho2020denoising}. The overwhelming success of diffusion models starts from image synthesis but extends to other modalities, like video, audio, text, graph, etc. Considering their dominant influence in the development of generative AI, we first summarize GAN and diffusion models before introducing other families of deep generative models. 

\begin{figure}[!htbp]
    \centering
    \includegraphics[width=\linewidth]{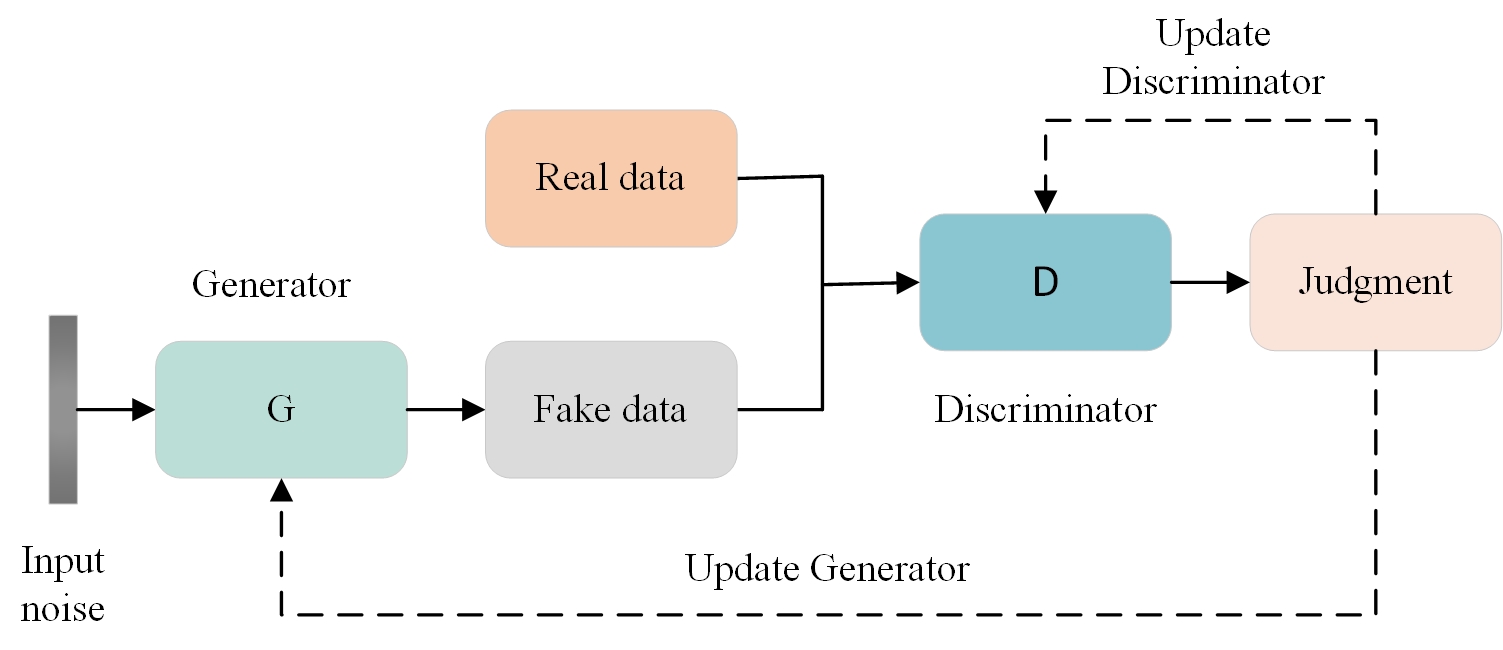}
    \caption{A schematic of GAN structure.}
    \label{fig:GAN}
\end{figure}

\textbf{GAN.}
The architecture of GAN is shown in Figure~\ref{fig:GAN}. GAN is featured by its two network components: a discriminator ($\mathcal{D}$) and a generator ($\mathcal{G}$). $\mathcal{D}$ distinguishes real images from those generated by $\mathcal{G}$, while $\mathcal{G}$ aims to fool $\mathcal{D}$. Given a latent variable $\bm{z} \sim p_{\bm{z}}$, the output of $\mathcal{G}$ is $\mathcal{G}(\bm{z})$ constituting a probability distribution $p_{\bm{g}}$. The goal of GAN is to make $p_{\bm{g}}$ approximate the observed data distribution $p_{\bm{data}}$. This objective is achieved through adversarial learning, which can be interpreted as a min-max game~\cite{schmidhuber1990making}:
\begin{equation} \label{eq:GAN-formula}
\begin{aligned}
	\min \limits_{G} \max\limits_{D}\hspace{2pt}&\mathbb{E}_{\bm{\mathrm{x}} \sim p_{data}} \mathrm{log}[D(\bm{\mathrm{x}})] + \mathbb{E}_{\bm{\mathrm{z}} \sim p_{\bm{\mathrm{z}}}} \mathrm{log}\left[1 - D(G(\bm{\mathrm{z}}))\right].
\end{aligned}
\end{equation}
where $\mathcal{D}$ is trained to maximize the probability of assigning correct labels to real images and generated ones, and is used to guide the optimization of $\mathcal{G}$ towards generating more real images. GANs have the weakness of potentially unstable training and less diversity in generation due to their adversarial training nature. The basic difference between GANs and autoregressive models is that GANs learn implicit data distribution, whereas the latter learns an explicit distribution governed by a prior imposed by model structure.

\begin{figure}[!htbp]
    \centering
    \includegraphics[width=\linewidth]{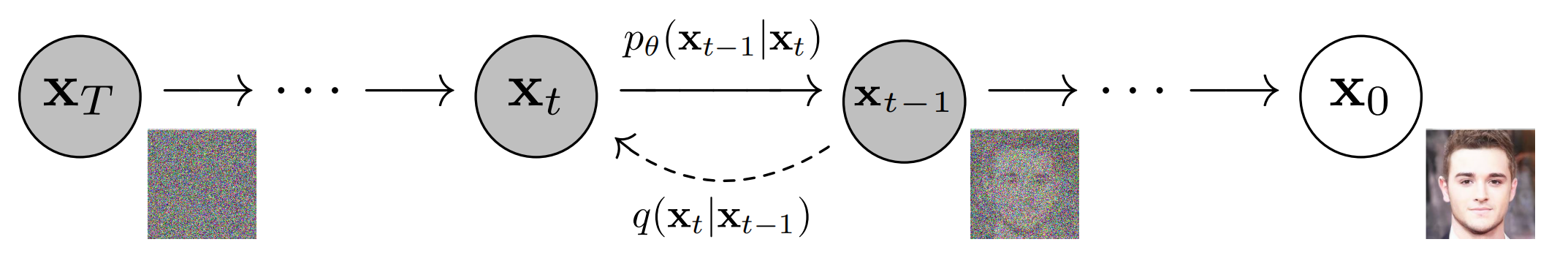}
    \caption{Diffusion model for image generation (figure obtained from \cite{ho2020denoising}).}
    \label{fig:DDPM}
\end{figure}

\textbf{Diffusion model.} The use of diffusion models, a special form of hierarchical VAEs,  has seen explosive growth in the past few years~\cite{ulhaq2022efficient, croitoru2022diffusion, cao2022survey, li2022diffusion, pascual2022full}. Diffusion models (Figure \ref{fig:DDPM}) are also known as denoising diffusion probabilistic models (DDPMs) or score-based generative models that generate new data similar to the data on which they are trained~\cite{ho2020denoising}. Inspired by non-equilibrium thermodynamics, DDPMs can be defined as a parameterized Markov chain of diffusion steps to slowly add random noise to the training data and learn to reverse the diffusion process to construct desired data samples from the pure noise.

In the forward diffusion process, DDPM destroys the training data through the successive addition of Gaussian noise. Given a data distribution $\mathbf{x}_0 \sim q(\mathbf{x}_0)$, DDPM maps the training data to noise by gradually perturbing the input data. This is formally achieved by a simple stochastic process that starts from a data sample and iteratively generates noisier samples $\mathbf{x}_T$ with  $q(\mathbf{x}_t\mid\mathbf{x}_{t-1})$, using a simple Gaussian diffusion kernel:
\begin{align}
q(x_{1:T} | x_0) := \prod_{t=1}^T q( x_t | x_{t-1} ), \label{eq:forwardprocess_1}
\end{align}
\begin{align}
q(x_t|x_{t-1}) := \mathcal{N}(x_t;\sqrt{1-\beta_t} x_{t-1},\beta_t I) \label{eq:forwardprocess_2}
\end{align}
where $T$ and  $\beta_t$ are the diffusion steps and hyper-parameters, respectively. We only discuss the case of Gaussian noise as transition kernels for simplicity, indicated as $\mathcal{N}$ in Eq.~\ref{eq:forwardprocess_2}. With  $\alpha_t := 1 - \beta_t$ and $\bar{\alpha}_t := \prod_{s=0}^{t} \alpha_s$, we can obtain noised image at arbitrary step $t$ as follows:
\begin{align}
q(x_t|x_{0}) := \mathcal{N}(x_t;\sqrt{\bar{\alpha}_t}x_{0},(1 - \bar{\alpha}_t) I) \label{eq:forwardprocess_3}
\end{align}

During the reverse denoising process, DDPM is learning to recover the data by reversing the noising process i.e., it undoes the forward diffusion by performing the iterative denoising. This process represents data synthesis and DDPM is trained to generate data by converting random noise into real data. It is also formally defined as a stochastic process, which iteratively denoises the input data starting from $p_\theta(T)$ and generates $p_\theta(x_0)$ which can follow the true data distribution $q(x_0)$. Therefore, the optimization objective of the model is as follows:
\begin{align}
     E_{t \sim \mathcal{U}( 1,T ), \mathbf x_0 \sim q(\mathbf x_0), \epsilon \sim \mathcal{N}(\mathbf{0},\mathbf{I})}{ \lambda(t)  \left\| \epsilon - \epsilon_\theta(\mathbf{x}_t, t) \right\|^2} \label{eq:loss}
\end{align}
Both the forward and reverse processes of DDPMs often use thousands of steps for gradual noise injection and during generation for denoising. 

\section{AIGC task: text generation} \label{sec:text}
NLP studies natural language with two fundamental tasks: understanding and generation. These two tasks are not exclusively separate because the generation of an appropriate text often depends on the understanding of some text inputs. For example, language models often transform a sequence of text into another, which constitutes the core task of text generation, including machine translation, text summarization, and dialogue systems. Beyond this, text generation evolves in two directions: controllability and multi-modality. The first direction aims to make the generated content

\subsection{Text to text}  
\subsubsection{Chatbots} 
\begin{figure}[b]
    \centering
    \includegraphics[width=\linewidth]{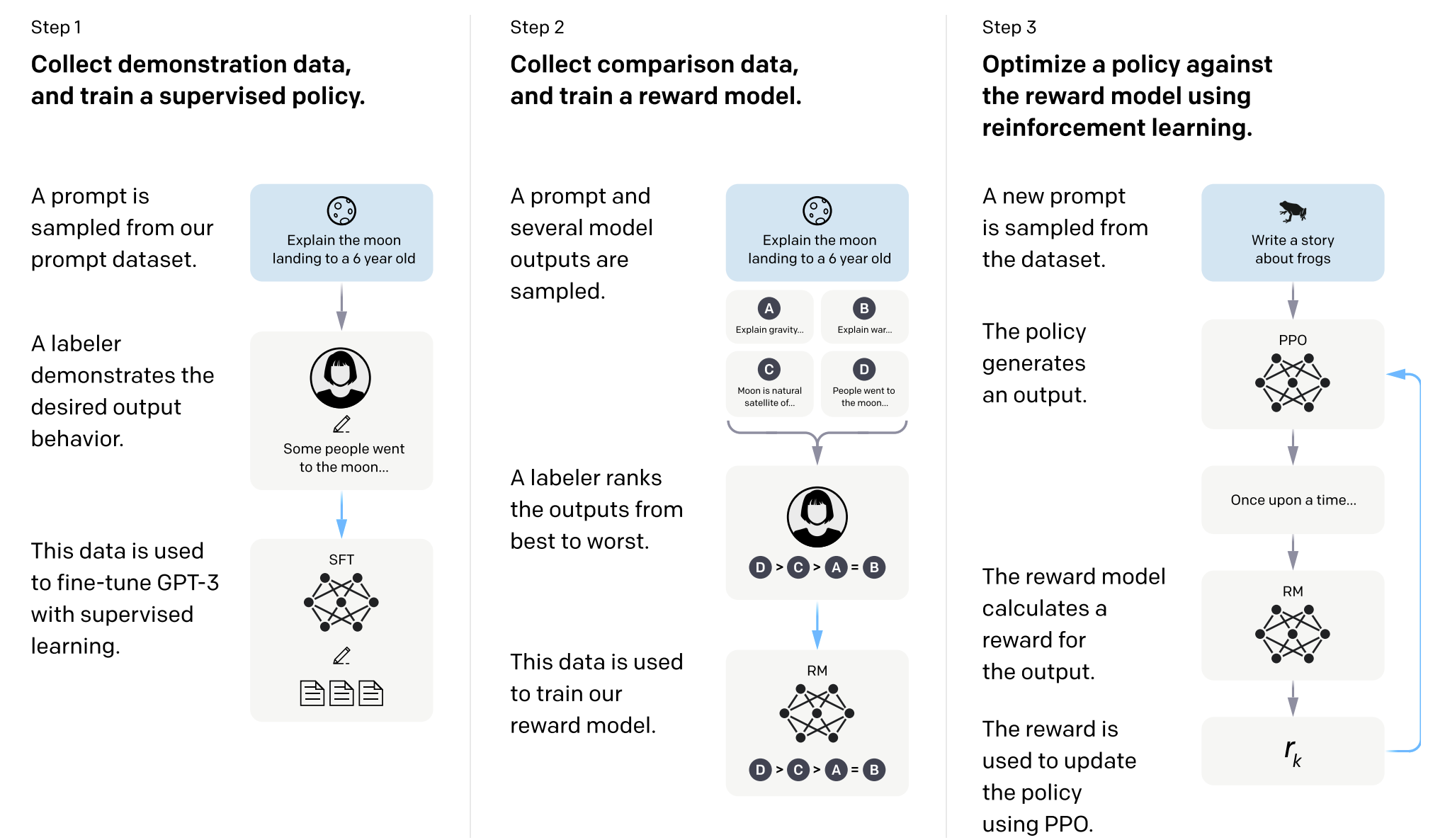}
    \caption{A diagram illustrating the three steps of how ChatGPT is trained by OpenAI (figure obtained from~\cite{ouyang2022training}).}
    \label{fig:chatbot}
\end{figure}

The main task of the dialogue system (chatbots) is to provide better communication between humans and machines \cite{ni2022recent, deriu2021survey}. According to whether the task is specified in the applications, dialogue system can be divided into two categories : (1) task-oriented dialogue systems (TOD)~\cite{zhang2020recent,peng2020few,yang2021ubar} and (2) open-domain dialogue systems (OOD)~\cite{zhou2020design,zhang2019dialogpt,adiwardana2020towards}. Specifically, the task-oriented dialogue systems focus on task completion and solve specific problems (e.g., restaurant reservations and ticket booking) ~\cite{zhang2020recent}. Meanwhile, open-domain dialogue systems are often data-driven and aim to chat with humans 
without task or domain restrictions~\cite{zhang2020recent,ritter2011data}.

\textbf{Task-oriented systems.} Task-oriented dialogue systems can be divided into modular  and end-to-end systems. 
The modular methods include four main parts: natural language understanding (NLU)~\cite{singla2020towards,su2019dual}, dialogue state tracking (DST)~\cite{shan2020contextual,wang2020slot}, dialogue policy learning (DPL)~\cite{huang2020semi,xu2020conversational}, and natural language generation (NLG)~\cite{baheti2020fluent,elder2020make}. After encoding the user inputs into semantic slots with NLU, DST, and DPL decide the next action that is then converted to natural language by NLG as the final response. These four modules aim to generate responses in a controllable way and can be optimized individually. However, some modules may not be differentiable, and the improvement of a single module may not lead to the improvement of the whole system~\cite{zhang2020recent}. To solve these problems, end-to-end methods either achieve an end-to-end training pipeline by making each module  differentiable~\cite{ham2020end,hosseini2020simple}, or use a single end-to-end module in the system~\cite{zhang2020probabilistic,yang2020graphdialog}. There still exist several challenges for both modular and end-to-end systems, including how to improve tracking efﬁciency for DST ~\cite{kim2019efficient,ouyang2020dialogue} and how to increase the response quality of end-to-end system with limited data~\cite{he2020amalgamating,henderson2019training,mehri2019pretraining}.

\textbf{Open-domain systems.} Open-domain systems aim to chat with users without task and domain restrictions~\cite{ritter2011data,zhang2020recent}, and can be categorized into three types: retrieval-based systems, generative systems,  and ensemble systems~\cite{zhang2020recent}. Specifically,  retrieval-based systems always find an existing response from a response corpus, while generative systems can generate responses that may  not appear in the training set. Ensemble systems combine retrieval-based  and generative methods by either choosing the best response or refining the retrieval-based model with generative one~\cite{zhang2020recent,zhu2018retrieval,serban2017deep}. Previous works improve the open-domain systems from multiple aspects, including dialogue context modeling~\cite{feng2020regularizing,jia2020multi,lin2020world,mehri2019pretraining}, improving the response coherence~\cite{xu2020conversational,gao2020dialogue,akama2020filtering,lison2017not} and diversity~\cite{qiu2019training,bao2019know,ko2020generating,su2020diversifying}. Most recently, ChatGPT (see Figure \ref{fig:chatbot}) has achieved unprecedented success and also falls into the scope of open-domain dialogue systems. Apart from answering various questions, ChatGPT can also be used for paper writing, code debugging, table generation, and to name but a few. 

\subsubsection{Machine translation} 
\begin{figure*}[!htbp]
    \centering
    \includegraphics[width=\linewidth]{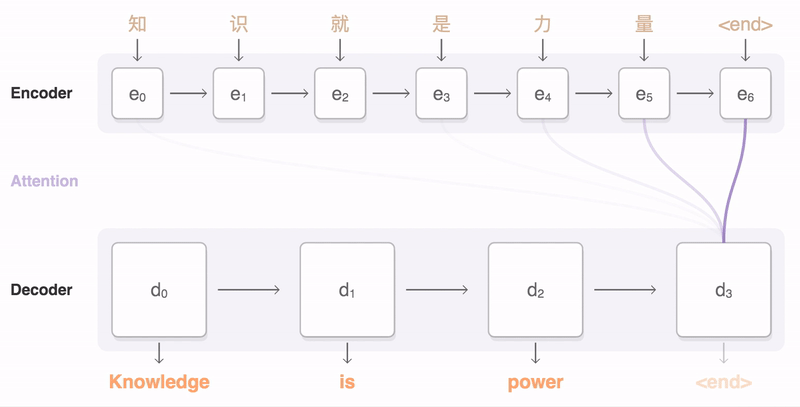}
    \caption{An example of machine translation (figure obtained from \cite{Britz:2017}).}
    \label{fig:machine translation}
\end{figure*}

As the term suggests, machine translation automatically translates the text from one language to another~\cite{hutchins1986machine,yang2020survey} (see Figure \ref{fig:machine translation}). With deep learning replacing rule-based~\cite{forcada2011apertium} and statistical~\cite{koehn2003statistical,koehn2007moses} methods, neural machine translation (NMT) requires minimum linguistic expertise~\cite{song1999general,wallach2006topic} and has become a mainstream approach featured by its higher capacity in capturing long dependency in the sentence~\cite{cho2014properties}. The success of neural machine learning can be mainly attributed to language models~\cite{bengio2000neural}, which predicts the probability of a word conditioned on previous ones. Seq2seq~\cite{sutskever2014sequence} is a pioneering work to apply encoder-decoder RNN structure~\cite{kalchbrenner2013recurrent} to machine translation. When the sentence gets long, the performance of Seq2seq~\cite{sutskever2014sequence} deteriorates, for which an attention mechanism was proposed in~\cite{bahdanau2014neural} to help translate the long sentence with additional word alignment. With increasing attention, in 2006, Google's NMT system helped reduce the translation effort of humans by around $60\%$ compared to Google’s phrase-based production system, which bridges the gap between Human and machine translation~\cite{wu2016google}. CNN-based architectures have also been investigated for NMT with numerous attempts~\cite{kaiser2016can,kalchbrenner2016neural}, but fail to achieve comparable performance as the RNN boosted by attention~\cite{bahdanau2014neural}. Convolutional Seq2seq~\cite{gehring2017convolutional} makes CNN compatible with the attention mechanism, showing CNN can achieve comparable or even better performance than RNN. However, this improvement was later outperformed by another architecture termed Transformer~\cite{vaswani2017attention}. With RNN or Transformer as the architecture, NMT often utilizes autoregressive generative model, where a greedy search only considers the word with the highest probability for predicting the next work during inference. 

A trend for NMT is to achieve satisfactory performance in low-resource setup, where the model is trained with limited bilingual corpus~\cite{wang2021survey}. One way to mitigate this data scarcity is to utilize auxiliary languages, like multilingual training with other language pairs~\cite{zoph2016multi,shatz2017native,johnson2017google} or pivot translation with English as the middle pivot language~\cite{ren2018triangular,cheng2019joint}. Another popular approach is to utilize pre-trained language models, like BERT~\cite{devlin2018bert} or GPT~\cite{radford2018improving}. For example, it is shown in~\cite{rothe2020leveraging} that initializing the model weights with BERT~\cite{devlin2018bert} or RoBERTa~\cite{liu2019roberta} significantly improves the English-German translation performance. Without the need for fine-tuning, GPT-family models~\cite{radford2018improving,radford2019language,brown2020language} also show competitive performance. Most recently, ChatGPT has shown its power in machine translation, performing competitively with commercial products (e.g., Google translate)~\cite{jiao2023chatgpt}.

\subsection{Multimodal text generation}
\subsubsection{Image-to-text}
\begin{figure}[!htbp]
    \centering
    \includegraphics[width=\linewidth]{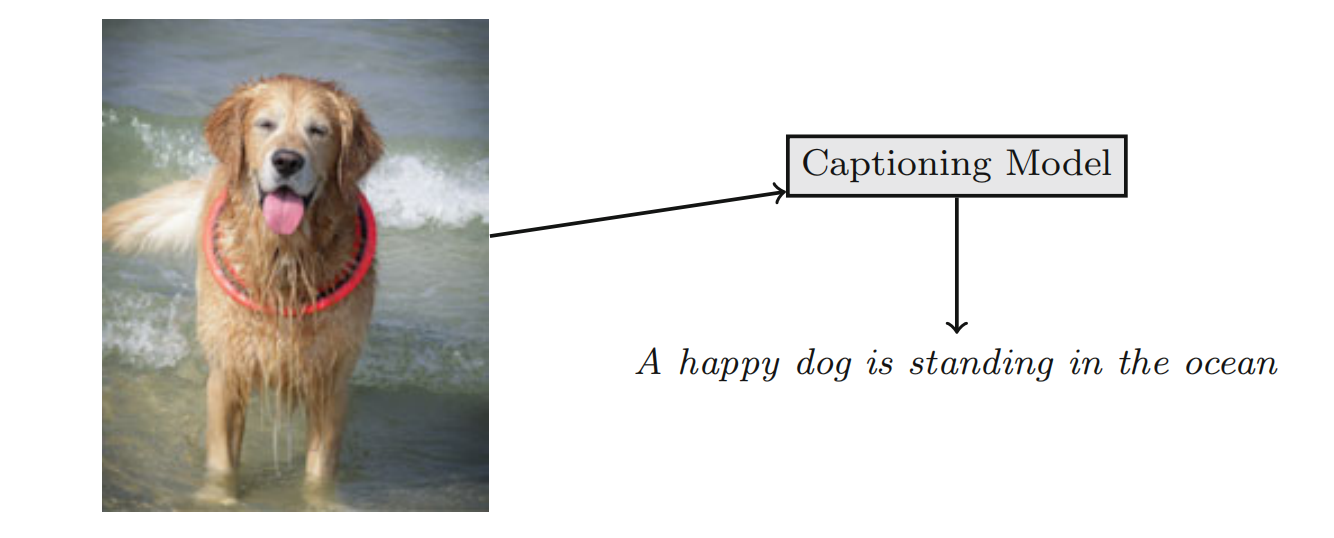}
    \caption{An example of image captioning (figure obtained from \cite{francis2021image}).}
    \label{fig:image_captioning}
\end{figure}

Image-to-text, also known as image captioning, refers to describing a given image's content in natural language (see Figure \ref{fig:image_captioning}). A seminal work in this area is Neural Image Caption (NIC)~\cite{vinyals2015show}, which employs CNN as an encoder to extract high-level representations of input images and then feed these representations into an RNN decoder to generate image descriptions. This two-step encoder-decoder architecture has been widely applied in later works on image captioning, and we term them as visual encoding~\cite{stefanini2021show} and language  decoding, respectively. Here, we first revisit the history and recent trends of both stages in image captioning.

\textbf{Visual encoding.} Extracting an effective representation of images is the main task of visual encoding module. Start from NIC ~\cite{vinyals2015show} with GoogleNet~\cite{szegedy2015going}  extracting the global feature of input image, multiple works adopt various CNN backbones as the encoder, including AlexNet~\cite{krizhevsky2012imagenet} in ~\cite{karpathy2015deep} and VGG network ~\cite{simonyan2014very} in ~\cite{mao2014deep,donahue2015long}. However, it is hard for a language model to generate fine-grained captions with global visual features. Following works introduce attention mechanism for fine-grained visual features, including attention over different grids of CNN features~\cite{xu2015show,lu2017knowing,wang2017skeleton,chen2018regularizing} or over different visual regions~\cite{anderson2018bottom,ke2019reflective,zha2019context}. Another branch of work~\cite{yang2019auto,zhao2021boosting} adopts graph neural networks to encode the semantic and spatial relationships between different regions. However, the human-defined graph structures may limit the interactions among elements~\cite{stefanini2021show}, which can be mitigated by the self-attention methods~\cite{yang2019learning,li2019entangled,zhang2021rstnet} (including ViT~\cite{liu2021cptr}) that connects all the elements.

\textbf{Language decoding.} In image captioning, a language decoder generates captions by predicting the probability of a given word sequence~\cite{stefanini2021show}. Inspired by the breakthroughs in the NLP area, the backbones of language decoders evolve from RNN~\cite{vinyals2015show,lu2017knowing,ke2019reflective,wang2020show} to Transformer~\cite{li2019entangled,herdade2019image,guo2020normalized}, achieving significant performance improvement. Beyond the visual encoder-language decoder architecture, a branch of work adopts BERT-like architecture that fuses the image and captions in the early stage of a single  model~\cite{li2020oscar,zhou2020unified,zhang2021vinvl}. For example, ~\cite{zhou2020unified} adopts a single encoder to learn a shared space for image and text, which is first pre-tained on large image-text corpus and finetuned, specifically for image captioning tasks.

\subsubsection{Speech-to-Text}
\begin{figure}[!htbp]
    \centering
    \includegraphics[width=\linewidth]{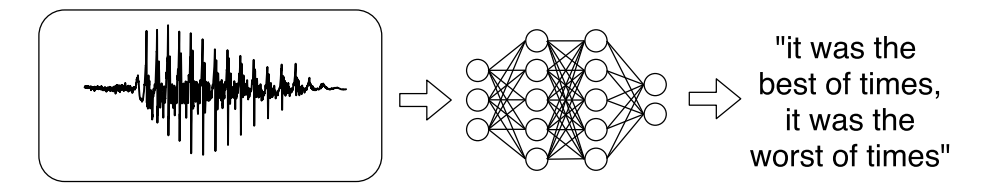}
    \caption{A example of speech recognition (figure obtained from \cite{carlini2018audio}).}
    \label{fig:ASR}
\end{figure}

Speech-to-text generation, also known as automatic speech recognition (ASR), is the process of converting spoken language, specifically a speech signal, into a corresponding text~\cite{reddy1976speech, indurkhya2010handbook} (see Figure~\ref{fig:ASR}). With many potential applications such as voice dialing, computer-assisted language learning, caption generation, and virtual assistants like Alexa and Siri, ASR has been an exciting field of research~\cite{raut2016automatic,karpagavalli2016review,malik2021automatic} since the 1950s, and evolved from hidden Markov models (HMM)~\cite{levinson1983introduction, juang1991hidden} to DNN-based systems~\cite{dahl2011context, hinton2012deep, graves2013speech, nassif2019speech, wu2020deep}.

\textbf{Various research topics and challenges.} Previous works improved ASR systems in various aspects. Multiple works discuss different feature extraction methods for speech signals~\cite{malik2021automatic}, including temporal features (e.g., discrete wavelet transform~\cite{milone2008learning,tang2009hybrid}) and spectral features such as the most commonly used mel-frequency cepstral coefficients (MFCC) ~\cite{toth2011hierarchical,collobert2016wav2letter,chiu2018state}. Another branch of work improves the system pipeline~\cite{roger2022deep} from multi-model~\cite{luscher2019rwth} to end-to-end ones~\cite{hori2018end,li2019jasper, nakatani2019improving, wang2019semantic, li2020comparison}. Specifically, a multi-model 
system~\cite{luscher2019rwth,malik2021automatic} first learns an acoustic model (e.g., a phoneme classifier 
that maps the features to phonemes) and then a language model for the word outputs~\cite{roger2022deep}. On the other hand, end-to-end models directly predict the transcriptions from the audio input~\cite{hori2018end,li2019jasper, nakatani2019improving, wang2019semantic, li2020comparison}. Although end-to-end models achieve impressive performance in various languages and dialects,  many challenges still exist. First, their applications for under-resourced speech tasks remain challenging as it is costly and time-consuming to acquire vast amounts of annotated training data~\cite{fendji2022automatic, roger2022deep}. Second, 
these systems may struggle to handle speech with specialized out-of-vocabulary words and may perform well on the training data but may not generalize well to new or unseen data~\cite{qin2013learning,fendji2022automatic}. Moreover, biases in the training data can also affect the performance of supervised ASR systems, leading to poor accuracy on certain groups of people or speech styles~\cite{benzeghiba2007automatic}.

\textbf{Under-resourced speech tasks.}  Researchers work on new technologies to overcome challenges in ASR systems, among which we mainly discuss the under-resourced speech problem that lacks data for impaired speech~\cite{roger2022deep}. A branch of work ~\cite{pascual2019learning,ravanelli2020multi}  
adopts multi-task learning to optimize a shared encoder for different tasks. Meanwhile, self-supervised ASR systems have recently become an active area of research without relying on a large number of labeled samples. Specifically, self-supervised ASR systems first pre-train a model on huge volumes of unlabeled speech data, then fine-tune it on a smaller set of labeled data to facilitate the efficiency of ASR systems. It can be applied for low-resource languages, handling different speaking styles or noise conditions, and transcribing multiple languages~\cite{liu2022audio, yadav2022survey, conneau2020unsupervised, baevski2019effectiveness}.

 \section{AIGC task: image generation} \label{sec:image}
Similar to text generation, the task of image synthesis can also be categorized into different classes based on its input control. Since the output is images, a straightforward type of control is images. Image-type control induces numerous tasks, like super-resolution, deblur,  editing, translation, etc. A limitation of image-type control is the lack of flexibility. By contrast, text-guided control enables the generation of any image content with any style at the free will of humans. Text-to-image falls into the category of cross-modal generation, since the input text is a different modality from the output image.

\subsection{Image-to-image}
\begin{figure}
    \centering
    \includegraphics[width=\linewidth]{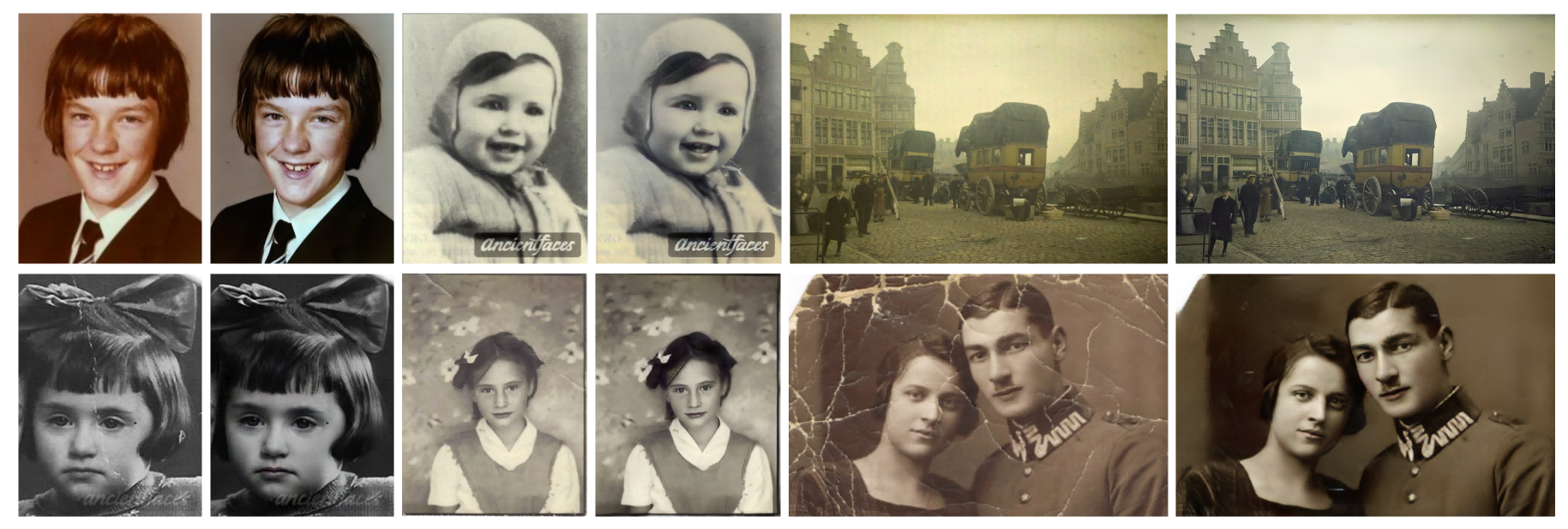}
    \caption{Examples of image restoration (figure obtained from \cite{wan2022old}).}
    \label{fig:img_restoration}
\end{figure}
\subsubsection{Image restoration} 
Image restoration solves a typical inverse problem that restores clean images from their corresponding degraded versions, with examples shown in Figure \ref{fig:img_restoration}. Such an inverse problem is non-trivial with its ill-posed nature because there are infinite possible mappings from the degraded image to the clean one. There are two sources of degradation: missing information from the original image and adding something undesirable to the clean image. The former type of degradation includes capturing a photo with a low resolution and thus losing some detailed information, cropping a certain region, and transforming a colorful image to its gray form. Restoration tasks recover them in order are image super-resolution, inpainting, and colorization, respectively. Another class of restoration tasks aims to remove undesirable perturbations, like denoise, derain, dehaze, deblur, etc. Early restoration techniques primarily use mathematical and statistical modeling to remove image degradations, including spatial filters for denoising~\cite{gonzalez2006woods, shrestha2014image, zhang2013gradient}, kernel estimation for deblurring~\cite{xu2010two, xu2016image}. Lately, deep learning-based methods ~\cite{jain2008natural, xie2012image, xu2014deep, liang2015stacked, dong2014learning,cheng2015deep,cai2020piigan, liu2021pd} have become predominant in image restoration tasks due to their versatility and superior visual quality over their traditional counterparts. CNN is widely used as the building block in image restoration~\cite{,wang2015deep, sun2015learning, dong2015image, varga2016fully}, while recent works explore more powerful transformer architecture and achieve impressive performance in various tasks, such as image super-resolution~\cite{liang2021swinir}, colorization~\cite{kumar2021colorization}, and inpainting~\cite{li2022mat}. There are also works that combine the strength of CNNs and Transformers together~\cite{fang2022hybrid, zhao2022rethinking,zhao2022hybrid}.

\textbf{Generative methods for restoration.} Typical image restoration models learn a mapping between the source (degraded) and target (clean) images with a reconstruction loss. Depending on the task, training data pairs can be generated by degrading  clean images with various perturbations, including resolution downsampling and grayscale transformation. To keep more high-frequency details and create more realistic images, generative models are widely used for restoration,  such as GAN in super-resolution~\cite{ledig2017photo, wang2018esrgan, zhang2019ranksrgan} and inpainting~\cite{nazeri2019edgeconnect, cai2020piigan, liu2021pd}. However, GAN-based models typically suffer from a complex training process and mode collapse. These drawbacks and the massive popularity of DMs led numerous recent works to adopt DMs for image restoration tasks \cite{Li2022SRDiffSI, saharia2022image, kawar2022denoising, saharia2022palette, lugmayr2022repaint, ren2022image}. Generative approaches like GAN and DM can also produce multiple variations of clean output from a single degraded image.

\textbf{From single-task to multi-task.} A majority of existing restoration approaches train separate models for different forms of image degradation. This limits their effectiveness in practical use cases where the images are corrupted by a combination of degradations. To address this, several studies~\cite{shin2022exploiting, kim2020restoring, zhou2022task, ahn2017image} introduce multi-distortion datasets that combine various forms of degradation with different intensities. Some studies~\cite{yu2018crafting, liu2019restoring, kim2020restoring, yuan2018unsupervised} propose restoration models in which different sub-networks are responsible for different degradations. Another line of work~\cite{shin2022exploiting, li2022all, zhou2022task, li2020learningdisentangled, suganuma2019attention} relies on attention modules or a guiding sub-network to assist the restoration network through different degradations, allowing a single network to handle multiple degradations. 

\subsubsection{Image editing}
\begin{figure}
    \centering
    \includegraphics[width=0.8\linewidth]{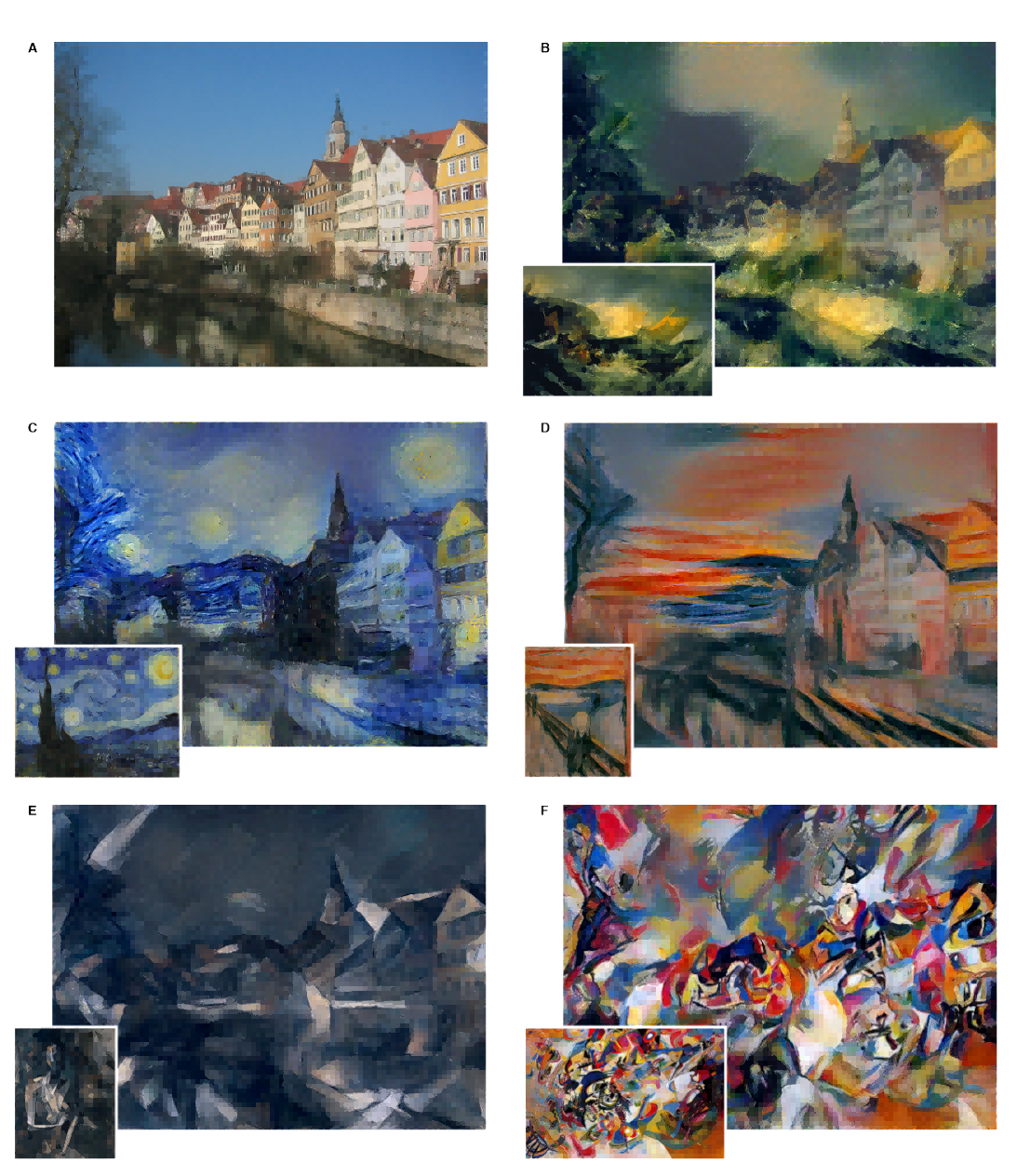}
    \caption{Examples of style transfer as a form of image editing (figure obtained from \cite{gatys2015neural}).}
    \label{fig:image editing}
\end{figure}

In contrast to image restoration for enhancing image quality, image editing refers to modifying an image to meet a certain need like style transfer (see Figure \ref{fig:image editing}). Technically, some image restoration tasks like colorization might also be perceived as image editing by perceiving adding color as the desired need. Modern cameras often have basic editing features such as sharpness adjustments~\cite{zhang2017optimized}, automatic cropping~\cite{zhang2005auto}, red eye removal~\cite{smolka2003towards}, etc. However, in AIGC, we are more interested in advanced image editing tasks that change the image semantics in various forms, such as content, style, object attributes, etc. 

A family of image editing targets to modify the attributes (like age) of the main object (like a face) in the image. A typical use case is facial attribute editing which can change the hairstyle, age, or even gender. Based on a pre-trained CNN encoder, a line of pioneering works adopt optimization-based approaches~\cite{li2016convolutional,upchurch2017deep}, which is time-consuming due to its iterative nature. Another line of works adopts learning-based approaches to directly generate the image, with a trend from single attribute~\cite{li2016deep,shen2017learning} to multiple ones~\cite{xiao2017dna,kim2017unsupervised,he2019attgan}. A drawback of most aforementioned methods is the dependence on annotated labels for attributes, therefore, unsupervised learning has been introduced to disentangle different attributes~\cite{shen2021closed,cherepkov2021navigating}.

Another family of image editing changes the semantics by combining two images. For example, image morphing~\cite{jing2019neural} interpolates the content of two images, while style transfer~\cite{gatys2016image} yields a new image with the content of one image and the style of the other. A naive method for image morphing is to perform interpolation in the pixel space, which causes obvious artifacts. By contrast, interpolating in the latent space can consider the view change and generate a smooth image. The latent space for those two images can be obtained via GAN inversion method~\cite{xia2022gan}. Numerous works~\cite{zhu2016generative,abdal2019image2stylegan,zhu2020domain,xu2021continuity} have explored the latent place of a pre-trained GAN for image morphing. For the task of style transfer, a specific style-based variant of GAN termed StyleGAN~\cite{Karras2019CVPR} is a popular choice. From the earlier layers to the latter ones, StyleGAN controls the attributes from coarser-grained (like structure) to  finer-grained ones (like texture). Therefore, StyleGAN can be used for style transfer by mixing the earlier layer's latent representation of the content image and the latter layer's latent representation of the style image~\cite{abdal2019image2stylegan,viazovetskyi2020stylegan2,guan2020collaborative,wei2022e2style}.

Compared with restoration tasks, various editing tasks enable a more flexible image generation. However, its diversity is still limited, which is alleviated by allowing other text as the input.  More recently, image editing based on diffusion models has been widely discussed and achieved impressive results~\cite{kim2021diffusionclip,chandramouli2022ldedit,hertz2022prompt,wallace2022edict}. DiffusionCLIP~\cite{kim2021diffusionclip} is a    pioneering work that finetunes a pre-trained  diffusion model to align the target image and text. By contrast, 
LDEdit~\cite{chandramouli2022ldedit}  avoids finetuning based on LDM~\cite{rombach2022high}. A branch of works discusses the mask problem in image editing, including how to connect a manually designed masked region and background seamlessly~\cite{avrahami2022blended1,avrahami2022blended1,avrahami2022blended2,ackermann2022high}. On the other hand, DiffEdit~\cite{couairon2022diffedit} proposes to predict the mask  automatically that indicates which part to be edited. There are also works editing 3D objects based on diffusion models and text guidance~\cite{li20223ddesigner,kim2022datid,chan2022efficient}. 

\subsection{Multimodal image generation}
\subsubsection{Text-to-image}
Text-to-image (T2I) task aims to generate images from textual descriptions (see Figure \ref{fig:text_to_image}.), and can be traced back to image generation from tags or attributes~\cite{srivastava2012multimodal,yan2016attribute2image}. AlignDRAW ~\cite{mansimov2015generating} is a pioneering work to generate images from natural language, and it is impressive that AlignDRAW~\cite{mansimov2015generating} can generate images from novel text like `a stop sign is flying in blue skies'. More recently, advances in text-to-image area can be categorized into three branches, including GAN-based methods, autoregressive methods, and diffusion-based methods.

\textbf{GAN-based methods.}   The limitation of AlignDRAW~\cite{mansimov2015generating} is that the generated images are unrealistic and require an additional GAN for post-processing. Based on a deep convolutional generative adversarial network (DC-GAN)~\cite{radford2015unsupervised}, ~\cite{reed2016generative} is the first end-to-end differential architecture from the character level to the pixel level. To generate high-resolution images while stabilizing the training process, StackGAN~\cite{zhang2017stackgan} and StackGAN++~\cite{zhang2018stackgan} propose a multi-stage mechanism that multiple generators produce images of different scales, and high-resolution image generation is  conditioned on the low-resolution images. Moreover, AttnGAN~\cite{xu2018attngan} and Controlgan~\cite{li2019controllable} adopt attention networks to obtain fine-grained control on the subregions according to relevant words.

\textbf{Autoregressive methods.} 
Inspired by the success of autoregressive Transformers~\cite{vaswani2017attention},  a branch of works generates images in an auto-regressive manner by mapping images to a sequence of tokens, among which DALL-E~\cite{ramesh2021zero} is a pioneering work.
Specifically, DALL-E~\cite{ramesh2021zero} first converts the images to image tokens with a pre-trained discrete variational autoencoder (dVAE), then trains an auto-regressive Transformer to learn the joint distribution of text and image tokens. A concurrent work CogView~\cite{ding2021cogview} independently proposes the same idea with DALL-E~\cite{ramesh2021zero} but achieves superior FID~\cite{heusel2017gans} than DALL-E~\cite{ramesh2021zero} on blurred MS COCO dataset. CogView2~\cite{ding2022cogview2}  extends CogView~\cite{ding2021cogview} to various tasks, e.g., image captioning,  by masking different tokens.
Parti~\cite{yu2022scaling} further improves the image quality by scaling the model size to 20 billion.

\begin{figure}[!htbp]
    \centering
    \includegraphics[width=0.8\linewidth]{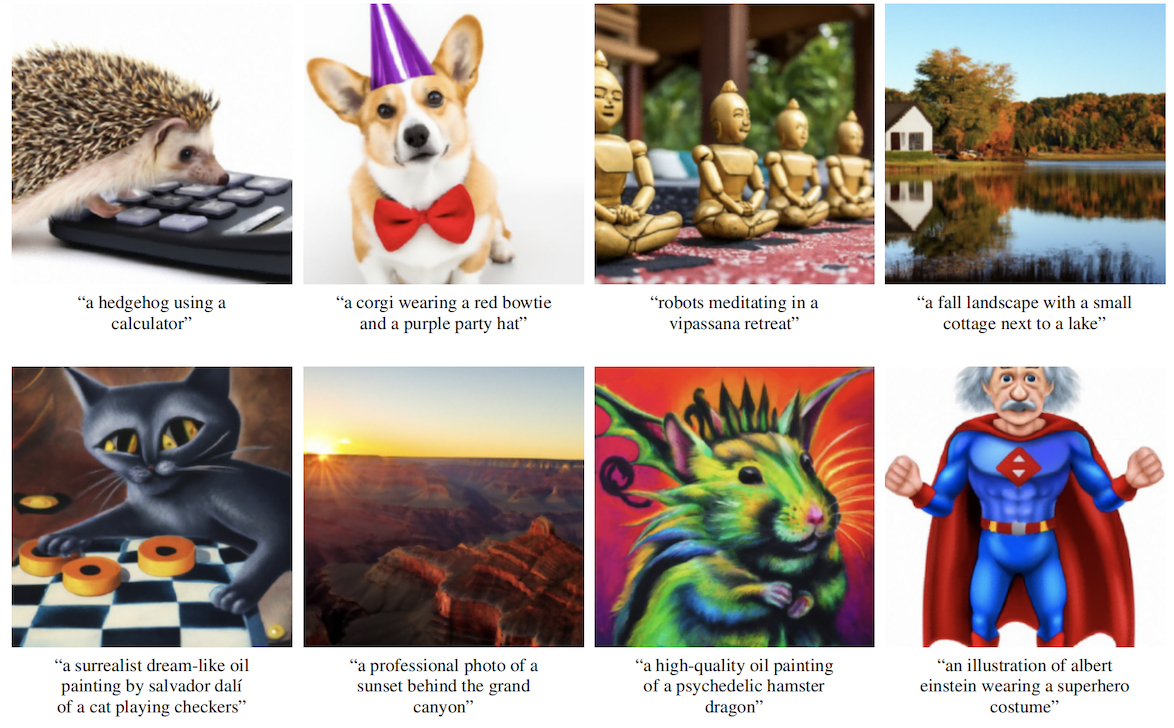}
    \caption{Examples of text-to-image (figure from~\cite{nichol2021glide}).}
    \label{fig:text to image}
\end{figure}

\textbf{Diffusion-based methods.} Diffusion model-based methods have achieved unprecedented success and attention recently, which can be categorized by 
either working on the pixel space directly~\cite{nichol2021glide,saharia2022photorealistic} or the latent space ~\cite{rombach2022high,ramesh2022hierarchical}. GLIDE~\cite{nichol2021glide} outperforms DALL-E by extending class-conditional diffusion models to text-conditional settings, while Imagen~\cite{saharia2022photorealistic} improves the image quality further with a pre-trained large language model (e.g., T5) capturing the text semantics.
 To reduce resource consumption of diffusion models in pixel space, Stable Diffusion~\cite{rombach2022high} 
 first compresses the high-resolution images to a low-dimensional  latent space,  then trains the diffusion model in the latent space. This method is also known as Latent Diffusion Models (LDM)~\cite{rombach2022high}. Different from Stable Diffusion~\cite{rombach2022high} that learns the latent space based on only images, DALL-E2~\cite{ramesh2022hierarchical} applies diffusion model to learn a prior as  alignment between image space and text space of CLIP. Other works also improve the model from multiple aspects, including introducing spatial control~\cite{avrahami2022spatext,voynov2022sketch} and reference images ~\cite{blattmann2022retrieval,sheynin2022knn}. 

\subsubsection{Talking face} \label{sec:talking}
\begin{figure}[b]
    \centering
    \includegraphics[width=\linewidth]{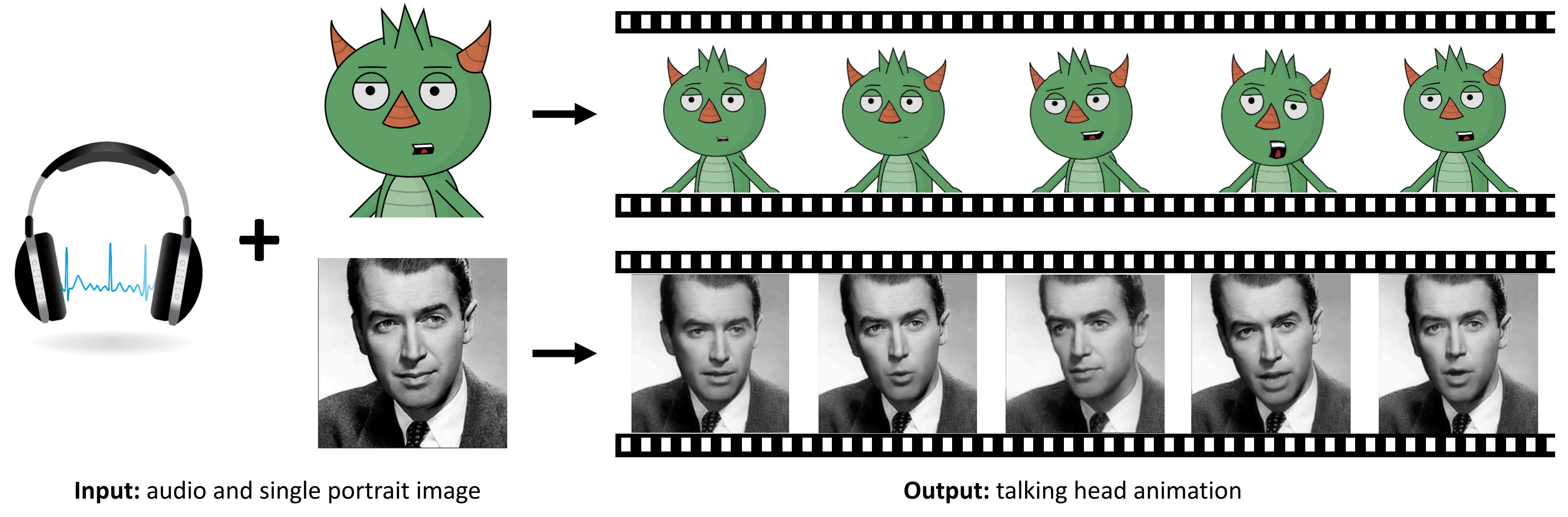}
    \caption{Examples of talking face (image obtained from \cite{chen2019hierarchical}).}
    \label{fig:talkingface}
\end{figure}

From the perspective of output, the task of talking face\cite{zhen2023human} generates a series of image frames which are thus technically a video (see Figure \ref{fig:talkingface}). Different from general video generation (see Sec.~\ref{sec:video}), talking face requires an image face as an identity reference, and edits it based on the speech input. In this sense, talking face is more related to image editing. Moreover, talking face converts a speech clip to a corresponding face image, resembling speech recognition to convert a speech clip to a corresponding word text. With speech recognition recognized as a multimodal generation text task, this survey considers talking face as a multimodal image generation task. 
Driven by deep learning models, speech-to-head video synthesis models have
attracted wide attention, which can be divided into 2D-based methods and 3D-based methods.

With 2D-based methods, talking face video synthesis mainly relies on landmarks, semantic maps, or similar representations. Landmarks are used as an intermediate layer from low-dimensional audio to high-dimensional video, as well as two decoders to decouple speech and speaker identity for generating video unaffected by speaker identity~\cite{chung2017you}, which is also the first work to use deep generative models to create speech faces. In addition, image-to-image translation generation~\cite{jamaludin2019you} can also be used for lip synthesis, while the combination of separate audio-visual representations and neural networks can also be used to optimize synthesis \cite{song2018talking, zhou2019talking}

Another line of work is based on building a 3D model and controlling the motion process through rendering technology \cite{suwajanakorn2017synthesizing, kumar2017obamanet}, with a drawback of high construction cost. Later, many generative talking face models based on 3DMM parameters \cite{karras2017audio, cudeiro2019capture, fried2019text, thies2020neural} were established, using models such as blendshape \cite{cudeiro2019capture}, flame \cite{li2017learning}, and 3D mesh \cite{richard2021meshtalk}, with audio as model input for content generation. At present, most methods are directly reconstructed from training videos. NeRF uses multi-layer perceptrons to simulate implicit representations, which can store 3D spatial coordinates and appearance information and are used for high-resolution scenes \cite{mildenhall2021nerf, muller2022instant, li2022nerfacc}. In addition, a pipeline and an end-to-end framework for unrestricted talking face video synthesis have also been proposed \cite{prajwal2020lip, kr2019towards}, taking any unidentified video and arbitrary speech as input.

\section{AIGC task: beyond text and image} \label{sec:others}

\subsection{Video} \label{sec:video}
Compared with image generation, the progress of video generation lags behind largely because of the complexity of modeling higher-dimensional video data. Video generation involves not only generating pixels but also ensuring semantic coherence between different frames. Video generation works can be categorized into unguided and guided generation (e.g., text, images, video, and action classes), with text-guided age (see Figure \ref{fig: video_generation}) receiving the most attention due to its high influence. 

\begin{figure}[!htbp]
    \centering
    \includegraphics[width=\linewidth]{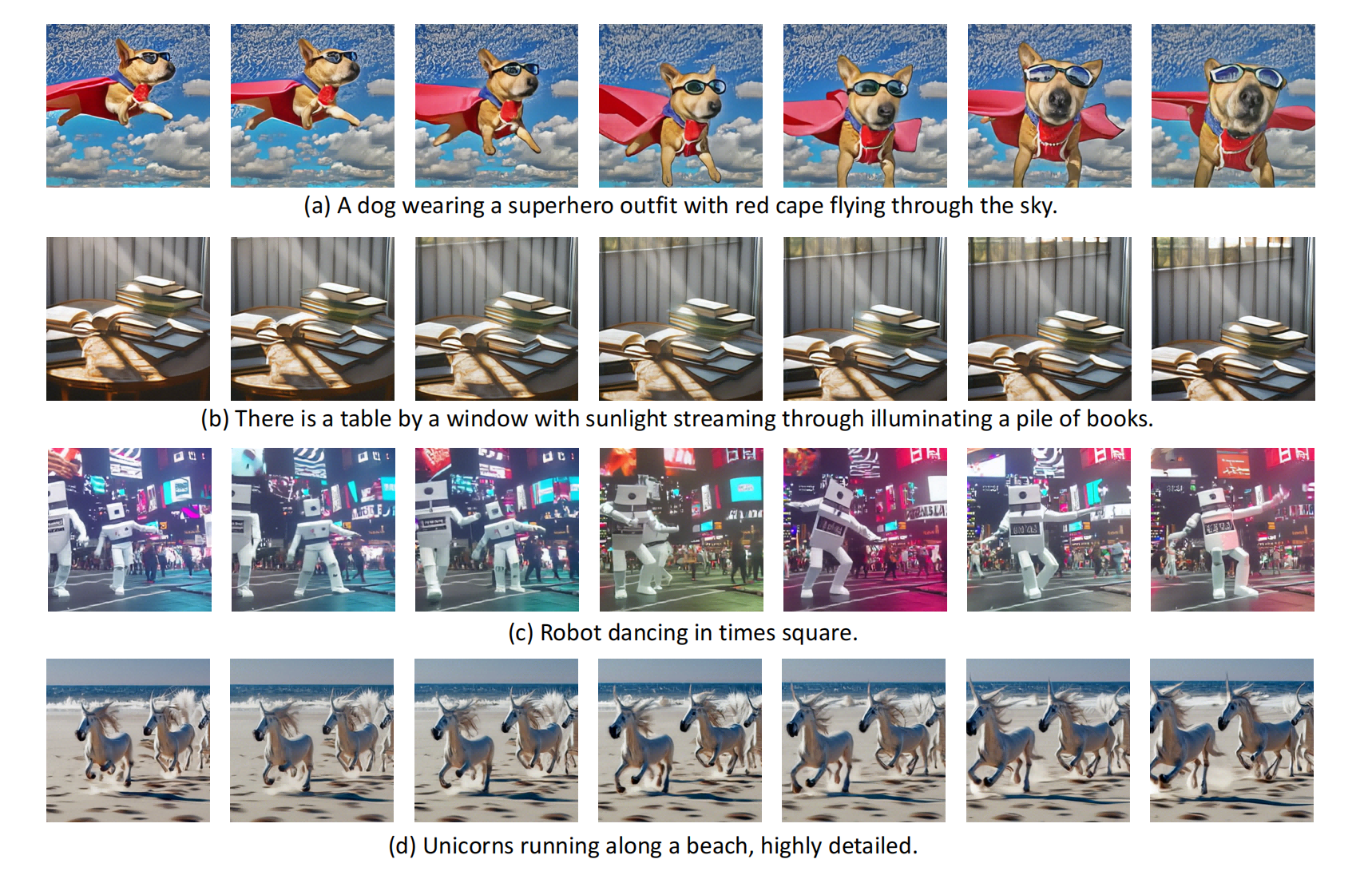}
    \caption{Examples of text-guided video generation (figure obtained from~\cite{singer2022make}).}
    \label{fig:video_generation}
\end{figure}

\textbf{Unguided video generation.} Early works on extending image generation from single frame to multiple frames are limited to creating monotonous yet regular content like sea waves. The generated dynamic textures~\cite{wei2000fast, doretto2003dynamic} often have a spatially repetitive pattern with time-varying visualization. 
With the development of generative models, numerous works~\cite{vondrick2016generating, 
saito2017temporal,ohnishi2018hierarchical,tulyakov2018mocogan, acharya2018towards, clark2019efficient, yushchenko2019markov} extend the exploration from naive dynamic textures to real video generation. Nonetheless, their success is limited to short videos for simple scenes with the availability of low-resolution datasets.
More recent works~\cite{clark2019adversarial, saito2020train, tian2021good, ho2022video} improve the video quality further,  among which~\cite{ho2022video} is regarded as a pioneering work of diffusion models.

\textbf{Text-guided video generation.} 
Compared to text-to-image models that can create almost photorealistic pictures, text-guided video generation is more challenging. Early works~\cite{mittal2017sync, pan2017create, marwah2017attentive, li2018video, gupta2018imagine, liu2019cross} based on VAE or GAN concentrate on creating video in simple settings, such as digit bouncing, and human walking. Given the great success of the VQ-VAE model in text-guided image generation, some works~\cite{wu2021godiva, hong2022cogvideo} extend it to text-guided video generation, resulting in more realistic video scenes. To achieve high-quality video, ~\cite{ho2022video} first applies the diffusion model to text-guided video generation, which refreshes the benchmarks of evaluation. After that, Meta and Google propose Make-a-Video~\cite{singer2022make} and Imagen Video~\cite{ho2022imagen} based on the diffusion model, respectively. Specifically, Make-a-Video extends a diffusion-based text-guided image generation model to video generation, which can speed up the generation and eliminate the need for paired text-video data in training. However, Make-a-Video requires a large-scale text-video dataset for fine-tuning, which results in a significant amount of computational resources. The latest Tune-a-Video~\cite{wu2022tune} proposes 
one-shot video generation, driven by text guidance and image inputs, where a single text-video pair is used to train an open-domain generator.

\subsection{3D generation}
The tremendous success of deep generative models on 2D images has prompted researchers to explore 3D data generation, which is actually a modeling of the real physical world.  Different from the single format of 2D data, a 3D object can be represented by  depth images, voxel grids\cite{wu20153d}, point clouds\cite{qi2017pointnet, qi2017pointnet++}, meshes\cite{hanocka2019meshcnn} and neural fields\cite{mescheder2019occupancy}, each of 
which has its advantages and disadvantages.


According to the type of input and guidance, 3D objects can be generated from text, images and 3D data. Although multiple methods \cite{jahan2021semantics, liu2022towards, fu2022shapecrafter} have explored shape editing guided by semantic tags or language descriptions, 3D generation is still challenging due to the lack of 3D data and  suitable architectures. Based on the diffusion model, DreamFusion~\cite{poole2022dreamfusion} proposes to solve these problems with a pre-trained text-to-2D model. Another branch of works  reconstruct the 3D objects from single-view images \cite{bednarik2018learning, tsoli2019patch, golyanik2018hdm, wang2018pixel2mesh, yuan2021vanet, li20213d} or multi-view images \cite{huang2018deepmvs, choy20163d, xie2019pix2vox, wang2021multi}, termed Image-to-3D. A new branch of multi-view 3D reconstruction is Neural Radiance Fields (NeRF) \cite{mildenhall2021nerf} for implicit representation of 3D information. The 3D-3D task includes  completion from partial 3D data \cite{wang2022pointattn} and transformation \cite{bai2015neural}, with 3D object retrieval as a representative transformation task.

\subsection{Speech}
Speech synthesis is an important research area in speech processing that aims to make machines generate natural and understandable speech from text. Methods of traditional speech synthesis include articulatory~\cite{kroger1992minimal, shadle2001prospects}, formant~\cite{allen1979mitalk, seeviour1976automatic}, concatenative synthesis~\cite{olive1977rule, moulines1990pitch}, and statistical parametric speech synthesis (SPSS)~\cite{kawahara2006straight, morise2016world}. 
These methods have been widely studied and applied, e.g., formant synthesis is still used in the open-source NVDA (one of the leading free screen readers for Windows). However, these generated speeches are identifiable from the human voice, and artifacts in synthesis speech reduce intelligibility.

Early works ~\cite{ze2013statistical,qian2014training, fan2014tts, zen2015acoustic, zen2015unidirectional} consist of three modules: text analysis, an acoustic model, and a vocoder. WaveNet~\cite{oord2016wavenet} is a revolution within speech synthesis which can generate the raw waveform from the linguistic features. To improve the quality of speech and diversity of voices,  generative models are introduced in speech synthesis, such as GAN~\cite{goodfellow2014generative}. Compared with GAN, diffusion models do not require a discriminator, making training more stable and simple. Therefore, the works of speech synthesis adopt diffusion models, becoming a rising trend. A branch of works~\cite{huang2022prodiff, xiao2021tackling,lam2022bddm, chen2022infergrad} focuses on efficient speech synthesis, in which different ways are adopted to reduce the generated time by accelerating inference, such as combining the schedule and score networks for training, jointly trained GAN. Another branch of study
~\cite{chen2021wavegrad,shi2022iton,mittal2021symbolic,rouard2021crash} concentrates on end-to-end models, which directly generate waveform from text without any intermediate representations. A fully end-to-end model not only simplifies the training and inference, but also reduces the demand for human annotations. The branch of diffusion-based speech synthesis is not limited to the two mentioned above, such as speech enhancement and guided speech synthesis.

\subsection{Graph}
Graphs are ubiquitous in the world, which aid in visualizing and defining the relationships between objects in a wide range of domains, from social networks to chemical compounds. 
Graph generation, which creates new graphs from a trained distribution that is similar to the existing graphs, has received a lot of attention.

Traditional graph generation works ~\cite{watts1998collective, albert2002statistical, leskovec2010kronecker} create new graphs with specific features that are related to the hand-crafted statistical features of real graphs
, which simplifies the process but fails to capture relational structure in complex scenarios. 
With the successes of deep learning algorithms, researchers have begun to apply them to graph generation, which, unlike the traditional methods, can be directly trained by real data and automatically extract features. Among them, works~\cite{you2018graphrnn, liao2019efficient, dai2020scalable} based on autoregressive model create graph structures sequentially in a step-wise fashion, which allows for greater scalability but fails to model the permutation invariance and is computationally expensive. Simultaneously, One-shot models~\cite{liu2018constrained, madhawa2019graphnvp, madhawa2019graphnvp} such as VAE and flow are incapable of accurately modeling structure information because of ensuring tractable likelihood computation. Although graph generation~\cite{de2018molgan, jin2018learning, maziarka2020mol} based on GAN sides step likelihood-based optimization by using a discriminator, the training is unstable.

Recently, there has been a surging interest in developing diffusion models for graph-structured data. EDP-GNN~\cite{niu2020permutation} is the pioneering to show the capability of the diffusion model in the Graph generation, with the goal of addressing non-invariant properties. After that, On the one hand, diffusion-based works~\cite{jo2022score, vignac2022digress, haefeli2022diffusion, luo2022fast, huang2022graphgdp} focus on realistic graph generation, which produces graphs that are similar to a given set of graphs. On the other hand,~\cite{shi2021learning, xie2021crystal, anand2022protein, zaman2022particlegrid} concentrate on goal-directed graph generation, which generates graphs that optimizes given objects, like molecular and material generation.

\subsection{Others}
There are also other interesting tasks generating content in different modalities, e.g.,  music generation~\cite{ji2020comprehensive} and lip-reading~\cite{fenghour2021deep}. A typical music generation system can be categorized into three representation levels (from top to bottom), which generates score, performance, and audio, respectively~\cite{ji2020comprehensive}. 
With the development of deep learning, music generation introduces various methods for higher music quality, e.g., MusicVAE~\cite{roberts2018hierarchical}, MuseGAN~\cite{dong2018musegan} and transformer in ~\cite{huang2020pop}. Music generation inspires and accelerates the development  of 
computer-assistant composition software, including Magenta project from Google and Flow Machine project from Sony Computer Science Laboratories. A Lip reading task transforms visual inputs of lip movement to decoded speech~\cite{fenghour2021deep}, and has also shown impressive advances thanks to improved corpora and architectures.

\section{Industry Applications} \label{sec:industry}
Undoubtedly, AIGC has gone viral on social media since 2022. For example, users are active in sharing their experience of using ChatGPT for having an interactive conversation or Stable diffusion for generating images with a text prompt. However, this hype is expected to dwindle if AIGC cannot be used for practical applications in the industry to demonstrate its value. Therefore, we discuss how AIGC might influence various industries.

\subsection{Education}
AIGC is changing the paradigm of education by assisting in teaching and learning. Generative AI carries transformative potential in teaching, with the application ranging from course materials generation to assessment and evaluation~\cite{zentner2022applied, pettinato2023chatgpt}. Simultaneously, applications of generative models have begun to influence how students learn~\cite{victor2021percent, baidoo2023education}.

Generative AI technologies can provide educators with creation of personalized tutoring~\cite{ zentner2022applied}, designing course materials~\cite{pettinato2023chatgpt}, and
assessment and evaluation~\cite{zentner2022applied, baidoo2023education}. A unique foreign language teaching product for young children using generative technologies such as ChatGPT can attract children's attention, motivate them, and provide a fun learning environment. Higher education needs to embrace the use of AI in higher education, which can create more engaging, effective, and efficient learning experiences for students \cite{zentner2022applied}. One of the primary benefits of generated AI course material generation is that it can save teachers time and effort by automating the process of creating and updating course material. In addition, ChatGPT could significantly reduce the workload of law school instructors, freeing up time to increase academic productivity or develop more complex teaching skills \cite{pettinato2023chatgpt}. The benefits of ChatGPT in promoting teaching include but are not limited to facilitating personalized and interactive learning. However, some limitations of ChatGPT, such as generating incorrect information, exacerbating existing biases in data training, and privacy issues, can also appear \cite{baidoo2023education}. Overall, addressing these challenges requires collaborative efforts from policymakers and educators to provide recommendations or guidance for the appropriate use of generative AI tools.

Moreover, generative AI technologies can help students write essays~\cite{victor2021percent}, at-home tests or quizzes~\cite{victor2021percent}, comprehend certain theories and concepts,  and different language essays and papers in academic issues~\cite{zentner2022applied, baidoo2023education}. Chatbots can provide students with 24/7 support, allowing them to get the help they need when they need it.  With the ability to correct grammar, suggest improvements, and identify weak areas, chatbots like ChatGPT can provide students with immediate feedback on their writing, helping them to learn from their mistakes and improve their writing skills over time. This not only saves students time but also helps them to become better writers~\cite{Karan2023education}.  According to a survey conducted by an online course provider, 89\% of students use ChatGPT to complete their homework, with 50\% using it for essays and 48\% using it for at-home tests or quizzes~\cite{victor2021percent}.
Additionally, generated AI can tailor the course material to individual students' needs, such as learning style and pace, which has the potential to improve student engagement and learning outcomes.
ChatGPT can also help students comprehend certain theories, concepts, and different language articles, making them work more effectively~\cite{zentner2022applied, baidoo2023education}. There are also challenges and concerns associated with generated AI course material generation, including the generated material's quality, and the possibility of bias in the data used to train the AI. As a result, before using generated course material in an educational context, it is critical to evaluate and validate it carefully~\cite{dawn2023content}.

With the use cases mentioned above, AIGC has the potential to revolutionize education by improving the quality and accessibility of educational content, increasing student engagement and retention, and providing personalized support for learners. With the continuous advancements in AI, AIGC is poised to become an integral part of the education industry, offering students a more engaging, accessible, and personalized learning experience.

\subsection{Game and metaverse}
Most users may not resonate with one-size-fits-all content in the game and metaverse, where personalization yields the best experience. Although games and metaverse provide users with virtual worlds, the content represents the character and personality of users. Generative AI makes that possible, which not only allows users to customize their avatars but also provides diverse scenarios and storylines, making the experience more immersive~\cite{ratican2023proposed, chen2022multimedia, plut2020generative, ratican2023proposed}. 

AI Dungeon powered by GPT-3 model allows users to generate an open-end story navigated by text, where generative AI will produce new events as the response to the different actions of users, creating a one-of-a-kind and unexpected gameplay experience~\cite{latitute2020ai}. Horizon Worlds, one of the most popular games, allows you to wander into the virtual world related to content consumption and creation. In Horizon, users will have more control over how they want to tailor their online experience to meet their individual needs. Specifically, users can design their unqiue avatars and scenes using gizmos that include pre-built object and avatar properties~\cite{meta2022what}. Moreover, the visual novel game Traveler concentrates on generating gorgeous scenarios to present users with visual impact, in which you will embark on a journey through a diverse world. When a player explores the game Traveler, the player will be exposed to magnificent visuals and immersive soundscapes. As each scene is unique, the content can range from dark forests to bustling cities, all crafted by generative AI~\cite{yocat2023traveler}. 

Although the term ``metaverse" has become a buzzword recently, in the real world, the virtual space created by the game may serve as the portal to the metaverse~\cite{ratican2023proposed, ning2021survey}. Roblox, a sandbox game platform, first included the concept of ``Metaverse" in its prospectus and made its market value soar, where players can create their own world beyond their imagination~\cite{ning2021survey}. Virtual concert singers have a more comprehensive range of musical styles and talents. Audiences can choose their own favorite styles and even idols, providing them with a more diverse and personalized concert experience. Travis Scott, a well-known American rapper and producer, performed a historic concert inside the Fortnite game, and his avatar guided the players to experience different scenes, ranging from underwater to outer space~\cite{andrew2020travis}. The University of California, Berkeley, presented its commencement ceremony in `Minecraft', a popular computer game. In the Minecraft game, students and alumni built a copy of campus using generative AI technology, allowing thousands of graduating seniors using their avatars from around the world to attend the event~\cite{joe2020thousands}.  
Overall, AI has played a significant role in the evolution of the game and metaverse, and its use continues to grow as technology improves and becomes more accessible.

\subsection{Media}
With the ubiquitous growth of generative AI technologies, they play a rising role in media and advertising. AIGC not only promotes the diversity of media, which provides a better experience for audiences, but it also enables media  practitioners more efficient in their work~\cite{miroshnichenko2018ai, latar2018robot, firat2019robot, zhang2022application,vakratsas2020artificial,campbell2022preparing,deng2019smart}.

The media powered with AIGC enables more diversified content and ways of reporting, changing the media  mode of production and organizational structure~\cite{zhang2022application, kim2017newspaper, marconi2020newsmakers}. AIGC can be applied to a variety of applications in media, such as writing robots, news anchors, and caption generation. Traditionally, media outlets have relied on expert journalists to write new articles and reports, which requires a significant amount of energy and time, resulting in a limited number of articles. Moreover, the timeliness of the news is critical, and the news may be eclipsed after an hour. Generative AI can greatly assist journalism by using text generation technologies to make journalism more efficient and responsive~\cite{miroshnichenko2018ai}. Associated Press applies these technologies to generate roughly 40000 stories a year and its articles on company earnings increase from 1200 to 14800~\cite{barbara2022facts}. The Quakebot, a robot reporter of Los Angeles Times News, only takes three minutes to complete a related article after the Los Angeles earthquake~\cite{will2014first}.
Bloomberg News, an international financial media company,  launched Buttetin in 2018, with the goal of providing personally storied whose one-sentence summaries are generated by chatbots~\cite{max2018bloomberg}. 

AI news anchors have emerged as a result of the deep integration of generative AI in the media~\cite{wang2021application, xue2022you, wang2021research }. AI news anchors, combined with real anchors, make the way to spread information more diverse. AI news anchors can broadcast news based on the text, whose appearance and expression imitate the real anchor. China's state news agency Xinhua and Chinese search engine, Sogou, have developed AI news anchors with different profiles and languages. The most impressive is the 3D AI news anchor Xin Xiaowei, whose broadcast form can be presented in all directions from various angles, significantly improving the sense of three-dimensionality and layering~\cite{alex2022china}. Additionally, Korea's cable channel MBN  has created the AI news anchor AI Kim, who can quickly respond to various emergencies and even report all day~\cite{alex2020china}. 
To help the hearing-impaired people to get more information about international sporting events and Beijing Winter Olympics, an AI-driven sign language service provider,
namely Ling Yu, was developed by giant tech Tencent, where AIGC tasks are used including 3D digital human modeling, machine translation, image generation, and speech-to-text~\cite{xia2022ai}. Moreover, Migu, a Chinese business dedicated to providing digital information, can offer smart subtitle functions to the live broadcast of Beijing Winter Olympics. Therefore, people with hearing impairments can watch the live broadcast of the sports event, which makes them more immersive. 

\subsection{Advertising}
Various AI applications have transformed the advertising industry, giving advertisers powerful tools to create innovative and engaging content that connects to consumers at a deeper level \cite{vakratsas2020artificial,deng2019smart,guo2021vinci}. Among the various applications of AI in advertising, AIGC is particularly influential by allowing advertisers to create personalized and attractive content that resonates with individual consumers. A creative advertising system (CAS) aligns with the principles of AI for generating and testing advertising ideas, which helps aspiring and mature creators understand that creativity is not an elite privilege, but rather a systematic process that can be assisted through data and computation \cite{vakratsas2020artificial}. Implementing programmatic advertising has not fully utilized self-generating technology, resulting in different consumers being exposed to the same content. Fortunately, a personalized advertising copy intelligent generation system (SGS-PAC) can automatically personalize advertising content to meet individual consumer needs \cite{deng2019smart}. Advertising posters are a common form of information display used to promote products. Another intelligent system, Vinci, supports the automatic generation of advertising posters \cite{guo2021vinci}. By inputting product images and slogans specified by users, Vinci uses deep-generation models to generate beautiful posters. 

In addition, Brandmark.io is an AIGC-based tool that automatically generates logos for businesses. The tool creates multiple logo variations based on the user's preferences and specifications. Advertisers can purchase and use the logos created by the tool for their businesses, making it an easy and cost-effective solution for logo design \cite{Shitanshu2023market}. By GAN that forces output to include specific keywords, the approach automates product listing generation likely to attract potential buyers \cite{martinez2018using}. It enhances users' marketing efforts on peer-to-peer marketplaces. Moreover, technological innovations have provided digital and automated tools to the advertising industry, but have also allowed advertisers to automate the production of ``synthetic advertising". As reported by \cite{shah2020research}, AIGC has transformed the advertising industry by enabling advertisers to create highly personalized and engaging content at scale while saving time and resources. We expect to see even more innovative and influential applications of generated AI in advertising.

\subsection{Movie}

It is interesting to see how technology now affects almost every step of movie creation. Research has led to the development of computer-based surroundings that help with editing, labeling, video retrieval, and many more~\cite{davis1994media, gordon1995conceptual, parkes1988artificial, parkes1989prototype, parkes1989settings, sack1994idic, swanberg1993architecture, tonomura1994structured, ueda1993automatic, nack1996auteur}. To start with, AI-powered screenwriting software has significantly impacted the movie-making process. AI has created a new movie experience by integrating visual effects (VFX)~\cite{momot2022artificial}, improved sound effects (SFX), and new viewing platforms. The 4K, IMAX, and 3D movies, as well as animations, are highly impacted by them.

The script forms the foundation of how a movie will fare at the ticket counters. AI-generation devices store and compute massive amounts of data to create ``ideal" scripts~\cite{anantrasirichai2022artificial}. AI software is also employed to rework old screenplays into polished versions that are then analyzed and improved by the director and writer. It goes beyond just developing and analyzing current scripts. Jasper AI~\cite{Jasper} and Scalenut~\cite{Scalenut} are two examples of AI scriptwriters.

Movies are given visual effects (VFX) to increase spectator appeal. They combine original images with real video to create engrossing, realistic, contextual depictions that may include digital surroundings, de-aging, and many more. The VFX team of The Curious Case of Benjamin Button~\cite{Benjamin} put up two arrays of cameras in a bright room and utilized the MOVA Contour reality capture technology~\cite{Mova} to construct a three-dimensional database of the hero's facial expressions. The VFX team next developed high-resolution 3D models of the lead character at various ages and lastly employed AI to manipulate the data retrieved from the three-dimensional database to cause the head models to age. The outcome was convincing and garnered widespread acclaim across the movie community.  In movies like Blade Runner 2049~\cite{BladeRunner} and Gemini Man~\cite{GeminiMan}, the VFX team tracked and recorded the protagonist's facial data with the help of motion capture technology~\cite{MotionCapture}. They then rebuilt the 3D facial data in a computer and further polished it to accomplish age deduction. Although this approach takes expensive technology and a vast amount of money, it is incredibly precise and adaptable.

AI is expanding the limits of amusement by bringing back actors who have passed away in movies. Deepfake technology uses computer visuals and AI to produce incredibly amazing and lifelike fake videos of actual people or made-up characters. Fast and Furious 7~\cite{Furious} used VFX and vintage video to bring back Paul Walker after he passed away in 2013, using the actor's visage transferred onto his sibling.

In addition to the visual effects, subtitles also play a vital role in viewers' experience. For the benefit of viewers having hearing impairments, automated subtitles for the deaf and hard of hearing, also known as SDH~\cite{Subtitles}, include textual transcriptions of speech, speaker changes, and background noise. These benefits substantially increase how much money movies make. Natural Language Processing (NLP), a kind of AI that focuses on deciphering spoken language, offers multilingual subtitles in movies. To produce automated subtitles, Rev~\cite{rev}, a cloud-based program, is widely used by movie lovers. AI has also revolutionized the task of speech prediction in silent movies. AI-generated speech synthesis can narrate silent movies and dub movies into multiple languages. Deep learning systems trained on massive human audio samples can produce natural-sounding voiceovers. Features like LPC (Linear Predictive Coding)~\cite{ephrat2017vid2speech} and mel spectrograms~\cite{ephrat2017improved} generate high-quality intelligible audio through conversion. Recently, a Tacotron2 model~\cite{shen2018natural} variation for the video-to-audio synthesis was proposed by~\cite{prajwal2020learning}.~\cite{yadav2021speech} suggests an efficient stochastic model that produces endless high-quality audio patterns for a specific silent video, thus effectively encapsulating the multimodality of the speech prediction issue. Along with the visual and sound effects, we have Colourlab.Ai~\cite{ColourlabAI} for color grading, Descript~\cite{Descript} for video editing, and many more tools continuously making waves in the movie industry.

\subsection{Music}

AIGC also makes it to the music industry with notable developments~\cite{yang2022comprehensive}. AI can not only spot patterns and trends in vast data sets that are challenging for humans to notice, but also allows amateur musicians a cutting-edge technique to enhance their creative process, which is a fantastic opportunity. The fusion of AI technology into music is a new trend that many experts, researchers, musicians, and record companies are exploring~\cite{bestmusic}. Many utilize AIGC to create entirely new music, while some software edit compositions in the style of various composers.

Music industries are anticipating significant expenditures in this field, whether it is because of using AI to compose music or to help musicians. A fantastic illustration of an AI melody generator is Google's Magenta project~\cite{hutson2017google},~\cite{alaeddine2021artificial}. IBM's Watson Beat is one more example. For composing an original song, it makes use of AI and machine learning~\cite{WatsonBeat},~\cite{frid2020music}. 2016 saw the successful creation of text-to-speech (TTS) recordings and recordings that resembled music by DeepMind researchers~\cite{DeepMind},~\cite{yang2017midinet}. AI is also vastly used for the processing and improvement of digital audio. LANDR, an incredible AI-powered creative tool that enables musicians to get their music on several streaming services like Spotify and Apple Music, is one such service. A significant problem known as ``writer's block~\cite{sexton2023film}" frequently confronts lyricists. But thanks to AI, it's no longer a problem now. Nowadays, many musicians employ AI to create new lyrics for their songs~\cite{LyricStudio}. GPT-2~\cite{GPT2},~\cite{zhou2023comprehensive}, a text-generating tool, has been created by OpenAI, an AI technology firm. Not only can this remarkable text generator produce authentic news, but it can also write lyrics for Beatles songs and music from all other genres. However, AI is not just capable of producing text; it can also create original soundtracks and melodies. The Sony CSL flow machine offers assistance to artists so they can develop original music based on their ideas~\cite{SonyCSL}. One of the most well-known AI tools for writing unique music is called AIVA~\cite{MusicIndustry}. To produce a unique track, the user first chooses a pre-set style and then modifies a variety of variables, such as the key, instrumentation, time signature, etc. AIVA can deconstruct all the intricate auditory information saved into discrete characteristics while reading hundreds of musical compositions by renowned musicians like Bach and Mozart~\cite{drott2021copyright}. These qualities may then be interpreted again and used to produce a completely new musical work. Apart from the tools mentioned above, there are so many other applications that made a significant impact on the music industry, such as the iOS-based tool Amadeus Code~\cite{songwriting}, the cloud-based platform Amper~\cite{Amper}, Ecrett Music~\cite{Ecrett}, etc.

\subsection{Painting}
From offering automatic painting tools to encouraging creative experimentation, AIGC is revolutionizing the painting industry in many ways. AI programs can analyze pictures to produce color schemes, patterns, and textures that can make artwork. The automatic drawing tools generated using these algorithms are able to apply these patterns and textures to produce distinctive and intricate works of art~\cite{zou2021stylized}. AI can also analyze a person's preferences, interests, and style to create customized artwork. Empowering artists to create art specifically suited to their preferences and interests can increase their appeal and value.

The artwork created by MidJourney under the title \textit{`Space Opera Theatre'} earned first place in the Colorado State Fair Art Competition~\cite{paul2022ai}, demonstrating the capability of AI painting tools to produce excellent pieces of art. Midjourney is an excellent AI image generator with comprehensive functions, which is used by many artists to generate inspiration. The creation of various art forms by generative AI, such as abstract painting generation~\cite{li2020abstract}, Chinese shanshui painting~\cite{zhou2019interactive}, and Chinese ink paintings~\cite{chung2022interactively}, undoubtedly promotes the advancement of painting. Moreover, AIGC can assist in conservation and restoration~\cite{yu2022artificial},~\cite{hu2022analysis}. AI algorithms are capable of analyzing and repairing ruined artwork. These algorithms make it simpler for conservators to return the artwork to its initial state by detecting and removing dust, scratches, and other flaws. 

For non-professionals unfamiliar with drawing or animation, AIGC is also very helpful because it enables them to produce high-quality visual effects. By adding additional constraints to the diffusion model, ControlNet~\cite{ControlNet} can increase the variability of the produced images. It can describe the generated images along with those other constraints of border drawing, depth information, Hough line map, normal map, and posture estimation. AIGC has also started a new era of collaborative artwork~\cite{chang2022between,bublitz2019collaborative}. AI algorithms can create collaborative paintings that involve multiple artists working together. These algorithms can analyze the styles of each artist and produce a unified style that incorporates elements from all of the artists' works.

\subsection{Code development}

Generative AI can contribute to the field of code development \cite{sun2022investigating,guo2022automated,guo2021towards,jahic2019software}, where AIGC can create code without the need for manual coding. The work by~\cite{sun2022investigating} explores the interpretability requirements of generative AI for code and demonstrates how human-centered approaches can drive the development of explainable AI (XAI) technologies in new domains. To improve testing efficiency and increase test coverage, it is particularly important to generate high-quality test cases automatically \cite{guo2022automated,guo2021towards}. In order to optimize the efficiency of data engineering, a novel software engineering approach based on neural networks for dataset augmentation can be designed \cite{jahic2019software}. One of the popular applications in AI-generated code is Github's Copilot, an AI tool jointly developed by GitHub and OpenAI. Users can automatically complete code through GitHub Copilot using software development tools \cite{Thomas2022}. Moreover, AI-generated technology can also assist in code refactoring, which improves existing code without changing its original functionality. This can shorten the time for developers to refactor and improve the quality of the code. A popular code refactoring tool is DeepCode \cite{Prathamesh2023}, an AI-supported code review tool that can inspect your code and provide suggestions for improvement. In addition, AIGC can also make an impact on the e-commerce and finance industries \cite{houde2020business}. E-commerce platforms such as Amazon, JD.com, and so on can use AI-powered customer service to provide shopping guide services to customers, thereby saving costs for enterprises. Financial companies can use virtual investment advisors to advise customers on securities account opening, financial investment, and other related services.

\subsection{Phone apps and features}
Numerous AIGC applications have emerged as fun-oriented mobile apps, typically in the form of image and video editing. Photoshop is traditionally a common tool for image editing, but manual work is time-consuming and can result in unnatural or unrealistic output. In addition, video editing involves analyzing each video clip and making editorial decisions based on both the audio and visual content. This process is time-consuming because the video is a time-based, dual-track medium that requires careful consideration of every frame. Fortunately, some work \cite{bar2022text2live,soe2021automation,argaw2022anatomy} has explored the utilization of AI technologies behind AIGC, to the image or video editing, making the applications in AIGC such as face swapping and digital avatar possible. 

Some popular applications based on face swapping are gaining widespread popularity on the Internet. This technology uses advanced AI technologies to analyze and swap people's faces with their favorite celebrities or anyone else in seconds, making it easier and faster to use compared to traditional PS technologies. VanceAI, Voila AI Artist and FaceAPP are leading figures, with FaceApp being recognized as the best facial photo editing App, winning numerous awards, and being downloaded by over 500 million users and counting \cite{Rose2021FaceApps}. Another popular application is voice-changing technology. This technology can adjust the pitch, timbre, speech rate, and other characteristics of the human voice to change the quality of the human voice. MagicMic \cite{MagicMic} and Voicemod \cite{voicemodWebsite} are two popular applications for real-time voice modification and soundboard operations, which people can use to change their voices for creating fun content, live streaming, or other purposes, enhancing the enjoyment of communication between people.

In addition, another technological trend is to transform individuals into virtual characters, thereby increasing entertainment value. virtual characters are digital avatars of people in a virtual world, they can be partial replicas of real people or even completely digital versions. Apple's first ``digital avatar" technology, Animoji, focuses mainly on generating preset cartoon and animal characters and does not support custom generation \cite{MemojiMaggie2022}. The second generation of ``digital avatar" technology represented by iPhone's Memoji and Xiaomi's Mimoji started to support personalized avatar customization, which offers a variety of options, starting from hairstyle, eyes, nose, dresses, etc \cite{MemojiNick2019}. This upgrade allows users to create an avatar that not only can track their facial movements, but also look like them. Besides that, the created avatars can also be posted as comments in WeChat or Facebook chats, giving users a more personal way to express themselves on social media. Since then, digital avatar technology has become one of the standard features of smartphones among various smartphone manufacturers.

\subsection{Other fields}
Beyond the above fields, AIGC is expected to have applications in more fields. For example, the design and development of a novel drug are complex, costly, and time-consuming. On average, it takes around \$3 billion and more than 10 years for a new drug to be accepted by the market~\cite{Regina2021}. This motivates using AIGC to accelerate the drug discovery process and reduce costs. In 2018 DeepMind created AlphaFold \cite{ruff2021alphafold}, which can accurately predict the structure of proteins and has been considered a milestone for drug discovery and fundamental biology research. Its updated version AlphaFold2 was released in 2020 and had higher accuracy than the former. ProteinMPNN\cite{dauparas2022robust}, designed by Justas Dauparas, can design protein sequences for specific tasks, generating entirely new proteins quickly in just a few seconds. Besides directly exploiting the generated content, AIGC can also help workers in various fields improve their efficiency. For example, in medical consultation, the patient can rely on chatbots for basic medical advice, while turning to the doctor only for more severe cases. In manufacturing design, it is possible to combine AIGC with the widely used computer-aided design system to minimize the repetitive effort so that the designer can focus on the more meaningful part.

\section{Challenges and outlook} \label{sec:society}
\subsection{Challenges}
Even though AIGC has shown remarkable success in generating realistic and diverse outputs across various domains, there are still numerous challenges in real-world applications. Except for requiring a large amount of training data and compute resources, we list some of the most significant challenges as follows.

\begin{enumerate}

\item \textbf{Lack of interpretability.} While AIGC models can yield impressive outputs, it remains challenging to understand how the model arrives at the outputs. This is especially a  concern when the model generates an undesirable output. Such a lack of interpretability makes it difficult to control the output. 

\item \textbf{Ethical and legal concerns.} The AIGC model is prone to data bias. For example, a language model mainly trained on the English text can be biased toward western culture. Copyright infringement and privacy violations are the underlying legal concerns that cannot be ignored. Moreover, the AIGC model also has the potential for malicious use. For example, students can exploit these tools to cheat on their essay assignments, for which AI content detectors are desired. AIGC models can also be used for distributing misleading content for political campaigns. 

\item \textbf{Domain-specific technical challenges.} At the current and in the near future, different domains require their unique AIGC models. Each domain is still faced with its unique challenges. For example, Stable Diffusion, a popular text-to-image AIGC tool,
 occasionally generates output that is far from what the user desires, such
as drawing humans as animals, one person as two people, etc. ChatBot, on the other hand, makes factual mistakes occasionally. 
\end{enumerate}

\subsection{Outlook}
Despite its unprecedented popularity, generative AI is still in its early stage. Here, we present how AIGC might evolve in the near future.

\begin{enumerate}
\item \textbf{More flexible control.} A major trend of AIGC tasks is to realize more flexible control. Taking image generation as an example, early GAN-based models can generate images of high quality, but with little control. The recent diffusion models trained on large text-image data enable control through text instruction. This facilitates the generation of images that better match the users' needs. Nonetheless, current text-to-image models still require more fine-grained control so that the images can be generated in a more flexible manner

\item \textbf{From pertaining to finetuning.} Currently, the development of AIGC models like ChatGPT focuses on the pretraining stage. The corresponding technology is relatively mature; however, how to fine-tune these foundation models for the downstream tasks is an under-explored field. Different from training a model from scratch, the goal of finetuning needs to trade-off between the foundation model's original general capability and its adaptation performance on the new task. 

\item \textbf{From big tech companies to startups.} At present, AIGC technology has mainly developed big tech companies, like Google and Meta. With the support of big tech companies, some startup companies have emerged to show high potentials, like OpenAI (supported by Microsoft) and DeepMind (supported by Google). With the focus transition from core technology development to applications, more startup companies are expected to emerge due to increasing demand.

\end{enumerate}

\begin{figure}[!htbp]
\centering
\includegraphics[width=0.49\textwidth]{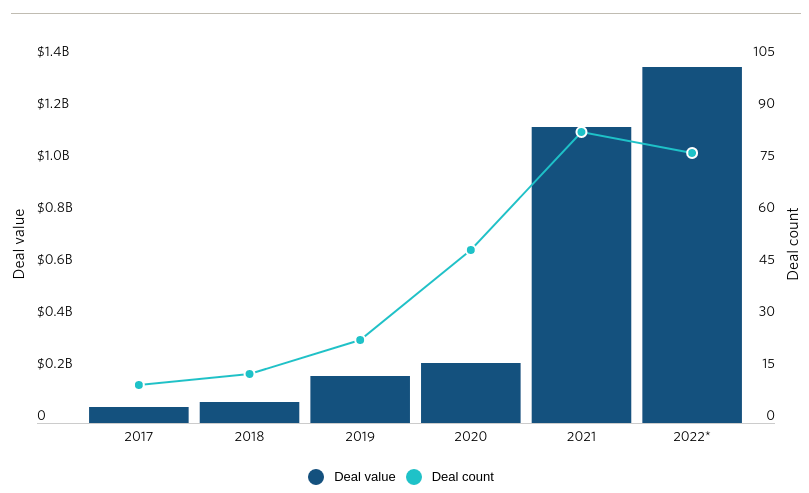}
\caption{Deal value and deal count for generative AI funding in the past 6 years (figure obtained from: \href{https://pitchbook.com/news/articles/generative-ai-venture-capital-investment}{PitchBook}).}
\label{fig:vc_invest}
\end{figure}

\textbf{Discussion: investment, bubble and job opportunities.} Technology-wise, there is no doubt that AIGC has made significant progress in the past few years. When a transformative technology emerges, the market tends to be over-optimistic about its potential applications and future growth, which also applies to generative AI. According to PitchBook (see Fig.~\ref{fig:vc_invest}), the funding for generative AI from venture capital (VC) increased significantly in the last two years. Some critics have concerns that generative AI might be the next bubble. One of their main concerns is that most AIGC tools are mainly playful instead of practical. For example, text-to-image models are fun to play with, but how they might generate revenues remains unclear. It is difficult to predict how generative AI might evolve. However, the authors of this work believe that generative AI is unlikely to become the next bubble considering it is a relatively new and rapidly growing field with many potential applications. There is also a hot debate about whether generative AI will replace humans, causing the loss of numerous job opportunities. On the other hand, generative AI can also create new job opportunities for individuals with skills on AI research and implementation skills. The industries that benefit from the power of AIGC might also boom and generate more job opportunities.